\documentclass{article}

\usepackage[preprint]{neurips_2023}

\usepackage[utf8]{inputenc} 
\usepackage[T1]{fontenc}    
\usepackage{hyperref}       
\usepackage{url}            
\usepackage{booktabs}       
\usepackage{amsfonts}       
\usepackage{nicefrac}       
\usepackage{microtype}      
\usepackage{xcolor}         

\usepackage{graphicx}
\usepackage{booktabs}
\usepackage{multirow}
\usepackage{colortbl}
\usepackage{amsmath}
\usepackage[capitalize]{cleveref}
\usepackage{natbib}

\definecolor{Gray}{gray}{0.9}

\setcitestyle{numbers}
\setcitestyle{square}
\setcitestyle{comma}

\newcommand{\pretraindataname}{DocStruct4M}
\newcommand{\modelname}{DocOwl 1.5}
\newcommand{\connectname}{H-Reducer}
\newcommand{\evalset}{DocLocal4K}
\newcommand{\instructset}{DocReason25K}

\title{mPLUG-DocOwl 1.5: Unified Structure Learning for OCR-free Document Understanding}

\author{%
  Anwen Hu\textsuperscript{1}, Haiyang Xu\textsuperscript{1}\thanks{Corresponding authors}, Jiabo Ye \textsuperscript{1}, Ming Yan\textsuperscript{1}$^*$ \\
  \textbf{Liang Zhang\textsuperscript{2}, Bo Zhang\textsuperscript{1}, Chen Li\textsuperscript{1}, Ji Zhang\textsuperscript{1}, Qin Jin\textsuperscript{2}, Fei Huang\textsuperscript{1}, Jingren Zhou\textsuperscript{1}}  \\
  \textsuperscript{1}Alibaba Group \\
  \textsuperscript{2}Renmin University of China \\
  \texttt{\{huanwen.haw,shuofeng.xhy,ym119608\}@alibaba-inc.com}}

\begin{document}

\maketitle

\begin{figure}[h]
\begin{center}
    \includegraphics[width=0.7\linewidth]{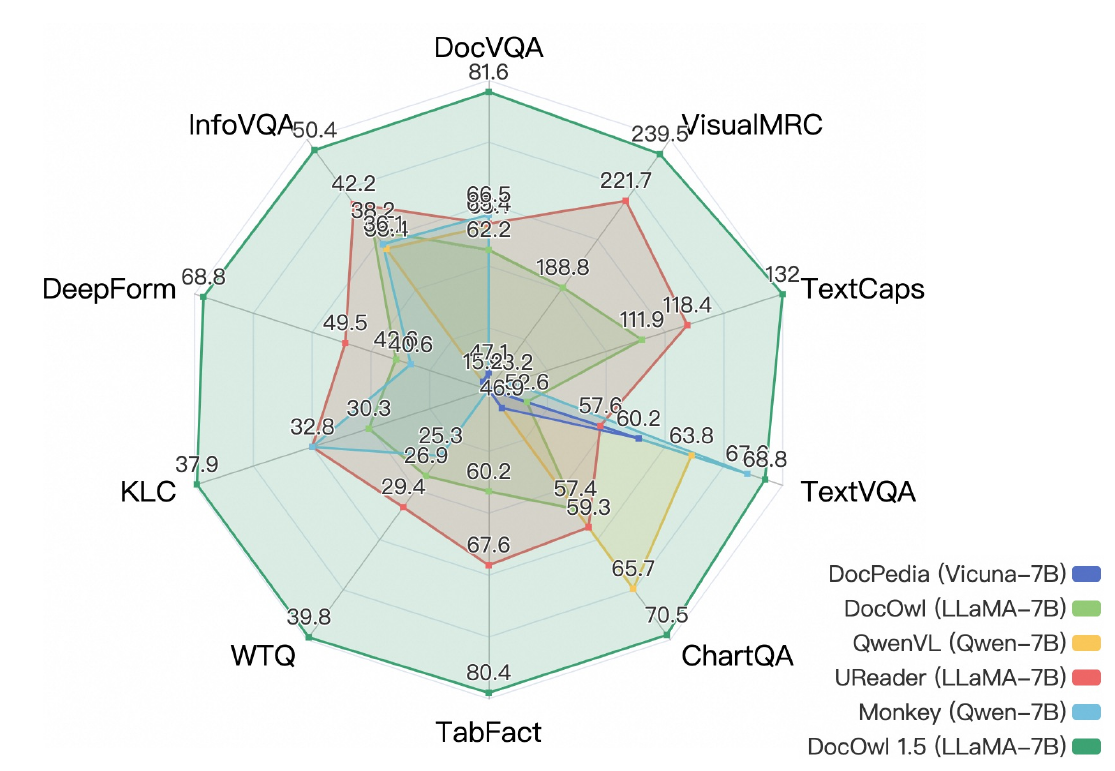}
\end{center}
\vspace{-8pt}
\caption{Compared with similar-size generalists, our \modelname~achieves state-of-the-art OCR-free performance on 10 Visual Document Understanding benchmarks.}
\label{fig:radar}
\end{figure}

\begin{abstract}
   Structure information is critical for understanding the semantics of text-rich images, such as documents, tables, and charts. Existing Multimodal Large Language Models (MLLMs) for Visual Document Understanding are equipped with text recognition ability but lack general structure understanding abilities for text-rich document images. In this work, we emphasize the importance of structure information in Visual Document Understanding and propose the Unified Structure Learning to boost the performance of MLLMs. Our Unified Structure Learning comprises structure-aware parsing tasks and multi-grained text localization tasks across 5 domains: document, webpage, table, chart, and natural image. To better encode structure information, we design a simple and effective vision-to-text module \connectname, which can not only maintain the layout information but also reduce the length of visual features by merging horizontal adjacent patches through convolution, enabling the LLM to understand high-resolution images more efficiently.
   Furthermore, by constructing structure-aware text sequences and multi-grained pairs of texts and bounding boxes for publicly available text-rich images, we build a comprehensive training set DocStruct4M to support structure learning.  Finally, we construct a small but high-quality reasoning tuning dataset \instructset~to trigger the detailed explanation ability in the document domain. Our model DocOwl 1.5 achieves state-of-the-art performance on 10 visual document understanding benchmarks, improving the SOTA performance of MLLMs with a 7B LLM by more than 10 points in 5/10 benchmarks. Our codes, models, and datasets are publicly available at \url{https://github.com/X-PLUG/mPLUG-DocOwl/tree/main/DocOwl1.5}.
  
\end{abstract}

\section{Introduction}
Leveraging the strong language understanding and generation ability of Large Language Models (LLM)~\citep{gpt3,llama,vicuna,llm_survey}, some recent works~\citep{mplugowl,mplug-owl2,llava,llava1.5,minigpt4,blip2} have developed Multimodal Large Language Models (MLLMs) for general vision-and-language understanding. By aligning a pre-trained visual encoder (e.g. the ViT/L-14~\citep{vit2021} from CLIP~\citep{clip}) and the LLM with a Vision-to-Text (V2T) module, these models present promising performance on understanding general images. However, they still face great challenges with images with rich text information, such as documents, webpages, tables, and charts~\citep{llmocr}. This is mainly because the visual encoder and V2T module are trained on general image-text pairs and not specifically optimized to represent the textual and structural information in text-rich images. 

\begin{figure*}[tp]
    \centering
    \includegraphics[width=1.0\linewidth]{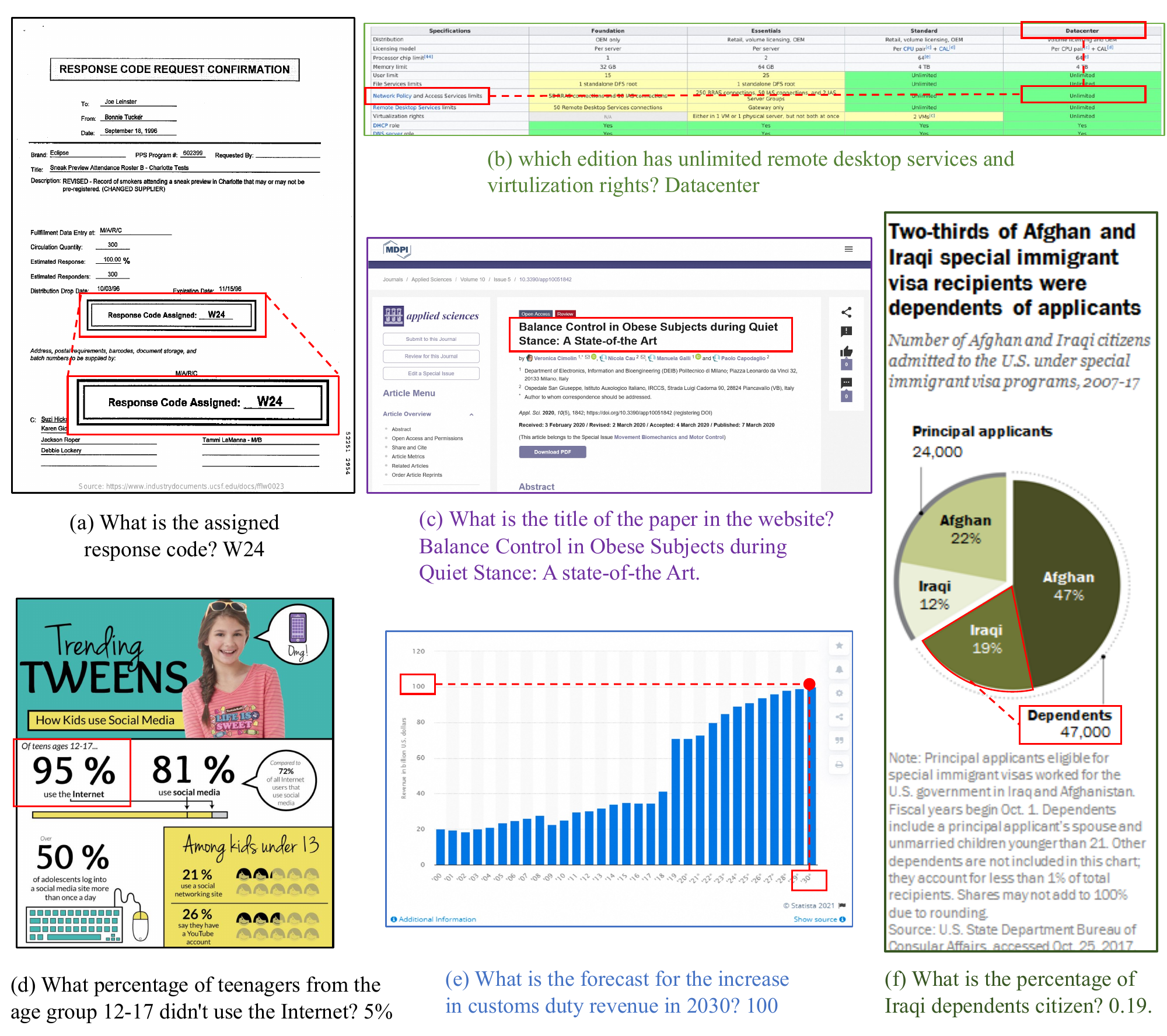}
    \caption{Illustrations of the importance of structure information in Visual Document Understanding on documents (a), tables (b), webpages (c), infographics (d), and charts (e-f).} 
    \label{fig:intro}
\end{figure*}

Textual information in images manifests with a multitude of visual structures, spanning the simplicity of plain text to the systematic grid layouts of tables and incorporating a spectrum of graphical representations such as pie, line, and bar charts. These elements may appear in isolation or be intricately interwoven within the framework of documents and webpages, reflecting a rich diversity of informational architecture across posters, invoices, infographics, scientific reports, academic and news websites, etc. As shown in \cref{fig:intro}, besides the basic textual content, structure information also plays a big role in Visual Document Understanding~\citep{layoutlmv2,layoutlmv3,udop,pix2struct}. With basic abilities to understand general images and comprehend structured texts through the LLM decoder, MLLM has the potential to achieve unified structure learning on text-rich images.
For better Visual Document Understanding with MLLMs, some works~\citep{docowl,ureader,qwenvl,docpedia} attempt to design text-reading tasks to strengthen the text recognition ability, but either ignore the structure comprehension or only cover limited domains of text-rich images, such as just webpages~\citep{pix2struct} or documents~\citep{docpedia}. In this work, we first propose to perform unified structure learning on text-rich images for MLLMs across 5 domains: document, webpage, table, chart, and natural image.

For better structural understanding, we first design a simple and effective vision-to-text module, namely \connectname. Unlike the Resampler~\citep{Alayrac2022FlamingoAV} or Q-former~\citep{blip2} which fuses visual features with learnable queries but affects spatial information, the \connectname~accumulates neighborhood visual features through convolution to keep the relative positional relationships. 
Compared with V2T modules with only linear layers~\citep{llava, llava1.5}, it produces much fewer visual features, which is more efficient for LLM to understand high-resolution document images. Considering texts in document images are most organized from left to right, \connectname~merges visual features at the horizontal level. Our Unified Structure Learning comprises structure-aware parsing tasks and multi-grained text localization tasks. To learn the organization of text contents, the former mainly teaches the model to parse the texts in the image in a structure-aware style, such as using line feeds and spaces to represent the structure of documents or webpages, and using extended Markdown syntax to represent the structure of tables and charts. Multi-grained text localization tasks further enhance the ability to correlate visually situated texts and concrete positions in the image. To support unified structure learning, based on publicly available datasets, we carefully build a comprehensive training set \pretraindataname~by constructing structure-aware sequences and multi-grained pairs of text and bounding boxes. The \modelname~is trained in a two-stage framework, starting with the Unified Structure Learning and then followed by the Multi-task Tuning among downstream tasks. Finally, to trigger the reasoning ability of MLLM in Visual Document Understanding, we construct a high-quality instruction tuning dataset~\instructset. By performing joint training on \instructset~and downstream datasets, \modelname-Chat well balance giving a simple answer or detailed explanations.

Our contributions in this work are four-fold:
\begin{itemize}
    \item We first propose Unified Structure Learning on text-rich images for MLLMs and design both structure-aware parsing tasks and multi-grained text localization tasks across 5 domains. A comprehensive dataset \pretraindataname~is carefully built to support Unified Structure Learning.
    \item We design a simple and effective vision-to-text module for structure learning and perform extensive experiments to validate its effectiveness.
    \item We construct a high-quality instruction tuning set to trigger the reasoning ability of MLLMs on Visual Document Understanding.
    \item \modelname~and \modelname-Chat achieves state-of-the-art OCR-free performance on 10 Visual Document Understanding tasks, achieving improvement of more than 10 points on 5/10 tasks among similar-sized models.
\end{itemize}

\section{Related Work}

\noindent{\textbf{Visual Document Understanding}}(VDU), also known as Visually-situated Language Understanding~\citep{pix2struct, ureader}, aims to comprehend images with rich text information. Such images range from documents~\citep{docvqa, infovqa,deepform,klc,mpmqa}, tables~\citep{wikitableqa,TabFact,pubtabnet}, charts~\citep{chartqa,dvqa,plotqa,chart2text,vistext, paperowl}, natural images~\citep{textcaps,textvqa,qctextcap} to webpage screenshots~\citep{visualmrc,websrc}, where diverse composition of text and visual objects contains a wealth of information. To evaluate the multimodal document understanding performance, the task formats include low-level recognition, e.g. information extraction~\citep{deepform, klc}, and high-level semantic understanding, such as visual question answering~\citep{docvqa, infovqa, wikitableqa, chartqa, visualmrc, textvqa}, image captioning~\citep{textcaps, chart2text, vistext}, and natural language inference~\citep{TabFact}. According to whether relying on an off-the-shelf OCR system to recognize texts in the image, models for Visual Document Understanding can be categorized into OCR-dependent models~\citep{udop,layoutlmv2,layoutlmv3,tap} and OCR-free ones~\citep{donut,pix2struct}. To leverage recognized texts from an OCR system, OCR-dependent models are always trained to align textual and visual inputs. For example, UDOP~\citep{udop} is pre-trained to recover masked text and layout information given image and retained text as inputs. As for OCR-free methods, training with tasks about text recognition is indispensable. Dount~\citep{donut} design the text reading task to output continuous text sequences that ignore structure information. 
To leverage structure information, Pix2Struct~\citep{pix2struct} designs a Screenshot Parsing Task to generate the HTML DOM tree for webpage screenshots but is hard to apply to other types of images. In this work, we first propose Unified Structure Learning for all image types and carefully build a comprehensive dataset to support layout learning.

\noindent{\textbf{Multimodal Large Language Models}}(MLLM) have shown strong vision understanding and open-ended conversation abilities~\citep{mplugowl,mplug-owl2, minigpt4, instructblip,qwenvl,cogagent,mmllm_survey} for natural images. They follow the architecture paradigm of connecting a vision encoder,e.g. ViT~\citep{vit2021,clip}, with a Large Language Model(LLM)~\citep{llama,vicuna,qwen} by a vision-to-text module, such as simple linear layers~\citep{llava,llava1.5} or a Q-Former~\citep{blip2}/Resampler~\citep{Alayrac2022FlamingoAV}/Abstractor~\citep{mplugowl, mplug-owl2} with learnable queries. To enable MLLMs to comprehend images with rich texts, there are major two challenges: how to encode high-resolution images and how to understand visually-situated texts. To tackle high-resolution images, most works choose to further train~\citep{qwenvl,docpedia} or extraly add a high-resolution vision encoder~\citep{cogagent}. UReader~\citep{ureader} first proposes to keep the low-resolution vision encoder and use a shape-adaptive cropping module to crop raw images into multiple sub-images with low resolution. To enhance the visually-situated text understanding, some work design tasks of reading texts from top-left to bottom-right without taking into account the importance of structure~\citep{ureader, qwenvl}. CogAgent~\citep{cogagent} and DocPedia~\citep{docpedia} further try strengthening the layout understanding for documents, webpages, and natural images with text grounding tasks. However, the comprehension of the overall structure is ignored, and tables and charts are not covered.  In this work, we follow UReader to process high-resolution images. To strengthen structure understanding, we design structure-aware praising and multi-grained text localization tasks for all types of images, covering documents, tables, charts, webpages, and natural images.  We propose a vision-to-text architecture to better maintain spatial information of visual features by convolution. Finally, to support unified structure learning, we build a comprehensive training dataset \pretraindataname~and greatly improve the visual document understanding performance.
\section{DocOwl 1.5}

\begin{figure*}
    \centering
    \includegraphics[width=1.0\linewidth]{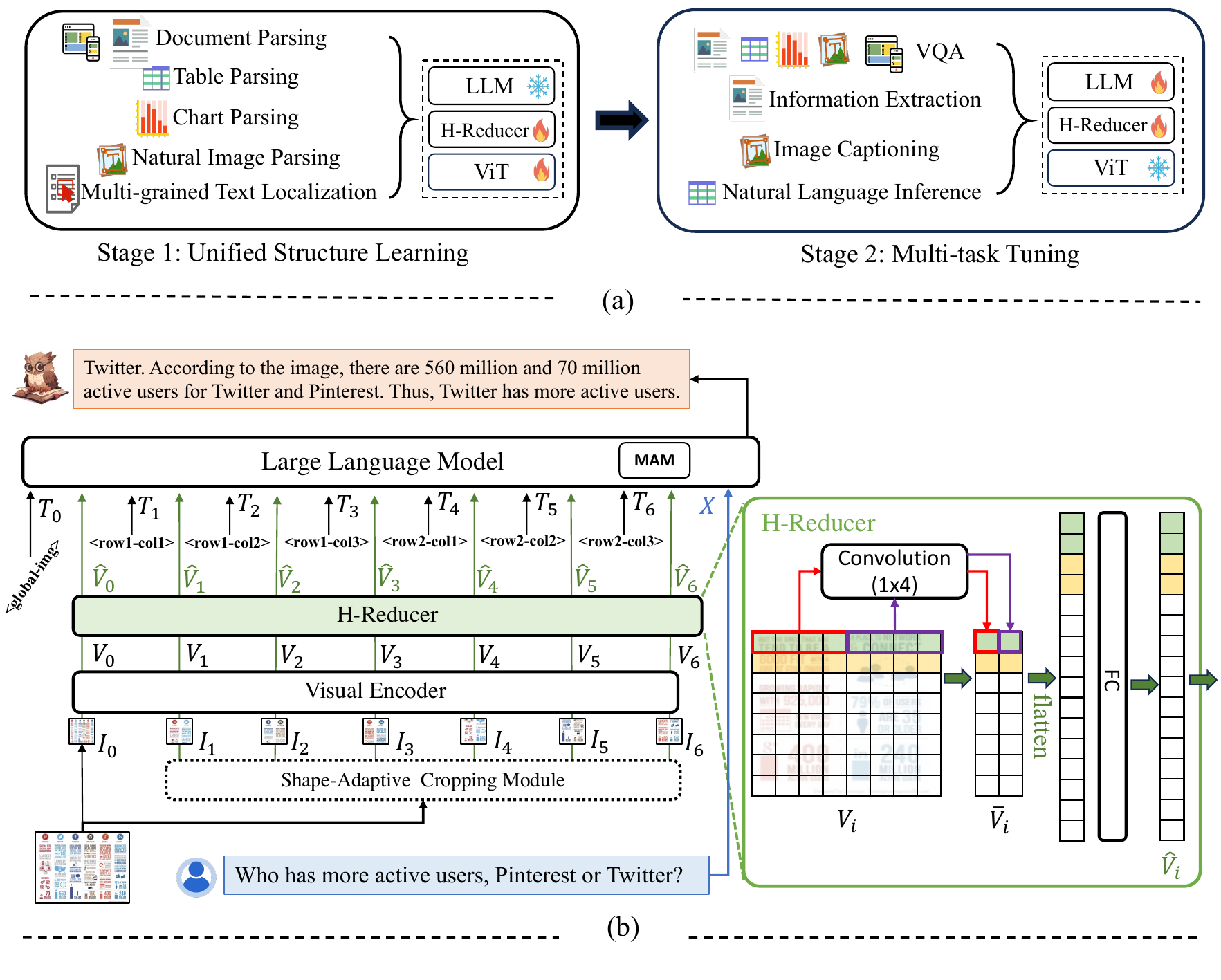}
    \caption{The two-stage training framework (a) and overall architecture (b) of \modelname. The global image and cropped images are processed independently by the Visual Encoder and \connectname. \texttt{<rowx-coly>} is the special textual token to indicate that the position of the cropped image in the original image is the $x^{th}$ row and $y^{th}$ column.} 
    \label{fig:overall_arch}
\end{figure*}

\modelname~follows the typical architecture of Multimodal Large Language Models, which consists of a visual encoder, a vision-to-text module, and a large language model as the decoder. To better keep the textual and layout information in text-rich images of high resolution, we design an \connectname~as the vision-to-text module to ensemble horizontal visual features. As shown in \cref{fig:overall_arch}(a), to enhance the text recognition and structure understanding abilities, we first perform Unified Structure Learning with structure-aware parsing and multi-grained text localization tasks for all types of images. Then, the model is jointly tuned on multiple downstream tasks of Visual Document understanding.

\subsection{Model Architecture}
\noindent\textbf{High-resolution Image Encoding.} As proved by previous works~\citep{donut,pix2struct,ureader}, the ability to encode high-resolution images is critical to ensuring that the decoder can use rich text information from document images. As shown in \cref{fig:overall_arch}(b), following UReader~\citep{ureader}
, we utilize a parameter-free Shape-adaptive Cropping Module to crop a shape-variable high-resolution image $I$ into multiple fixed-size sub-images $(I_1, I_2,...,I_C)$, where $C$ is the number of crops. To keep the overall layout information, the raw image is also resized to a low-resolution one as the global image $I_0$. Then, each image $I_i$ in $(I_0,I_1,...,I_C)$ is independently encoded to a sequence of visual features $V_i = (v_i^1, v_i^2,...,v_i^L), 0 \leq i \leq C$ by a transformer-based Visual Encoder, where $v_i^j, 1 \leq j \leq L$ is a $D$-dimension vector, $L$ is the length of visual features for each image.

\noindent\textbf{Spatial-aware Vision-to-Text Module: \connectname.} There are two kinds of popular vision-to-text modules for Multimodal Large Language Models: a MLP~\citep{llava, llava1.5,minigpt4} or a cross-attention module with learnable queries~\citep{mplugowl,qwenvl,Alayrac2022FlamingoAV,blip2}.  
Both two are not quite suitable for representing high-resolution text-rich images.
The former projects complete visual features into the language embedding space. It maintains all spatial information in the document image but keeps the sequence length of raw visual features, which is too long when processing high-resolution images. For example, encoding a 1,344x1,344 image with the ViT/L-14 results in 9,216 visual tokens. The cross-attention module could greatly reduce the length of the visual sequence to the number of learnable queries, but may lose spatial information during semantic fusion.

In this work, we design a more appropriate vision-to-text module for Visual Document Understanding, namely \connectname, which not only reduces visual sequence length but also keeps the spatial information. As shown in \cref{fig:overall_arch}(b), the \connectname~is comprised of a convolution layer to reduce sequence length and a fully-connected layer to project visual features to language embedding space. Since most textual information in document images is arranged from left to right, the horizontal text information is usually semantically coherent. Thus, the kernel size and stride size in the convolution layer are set as 1x4 to ensemble horizontal 4 visual features. The output channel is set equal to the input channel $D$. The convolution calculation is as follows:
 \begin{gather}
    V_i = (v_i^1, v_i^2,...,v_i^L)\\
    \overline{v}_i^j = f(v_i^{4j-3},v_i^{4j-2},v_i^{4j-1},v_i^{4j}), 1 \leq j \leq L/4, \\
    \overline{V}_i = (\overline{v}_i^1, \overline{v}_i^2,...,\overline{v}_i^{L/4}),
\end{gather}
where $f$ represents the dot product with kernel weights on multiple channels. After the convolution layer, the visual features of image $I_i$ are converted to the $\overline{V}_i$, the feature length of which is $L/4$.

Then, with a fully connected layer to align visual features to the language embedding space, the $\overline{V}_i$ are transferred to  $\hat{V}_i = (\hat{v}_i^1, \hat{v}_i^2,...,\hat{v}_i^{L/4})$.

\noindent\textbf{Multimodal Modeling with LLM.} As the decoder of MLLM, large language models should understand both the visual features of images and the textual features of language instructions. Following mPLUG-Owl2~\citep{mplug-owl2}, we apply the Modality-adaptive Module(MAM) in LLM to better distinguish visual and textual inputs. During self-attention, MAM utilizes two sets of linear projection layers to separately perform the key/value projection for visual features and textual features.
To help the LLM correlate multiple cropped sub-images, UReader~\citep{ureader} designs learnable crop position embeddings to denote the row and column position in the raw image. In this work, we simply add special textual tokens \texttt{`<row$x$\_col$y$>'} before the visual features of each cropped image, where $x$ and $y$ refer to the row and column index respectively. For the global image, the textual indicator token is \texttt{`<global\_img>'}.
This design eliminates the need to introduce additional parameters and is more friendly to the LLM decoder. Our experiments validate that it achieves comparable effects as the crop position embedding. Overall, the decoding of the LLM is as follows:
\begin{gather}
    Y = \rm{LLM}([T_0;\hat{V}_0, T_1;\hat{V}_1, ...,T_C; \hat{V}_C;X])
\end{gather}
where $[;]$ means the concatenation operation, $C$ is the crop number of the image, $T_j, 0 \leq j \leq C$ is the textual embeddings of the special textual indicator for the global image or positions of cropped images,  $\hat{V}_j$ is the visual features of a global or cropped image, $X$ is the textual embeddings of the instruction, $Y$ is the predicted answer.

\subsection{Unified Structure Learning}
\begin{figure*}[tp]
    \centering
    \includegraphics[width=1.0\linewidth]{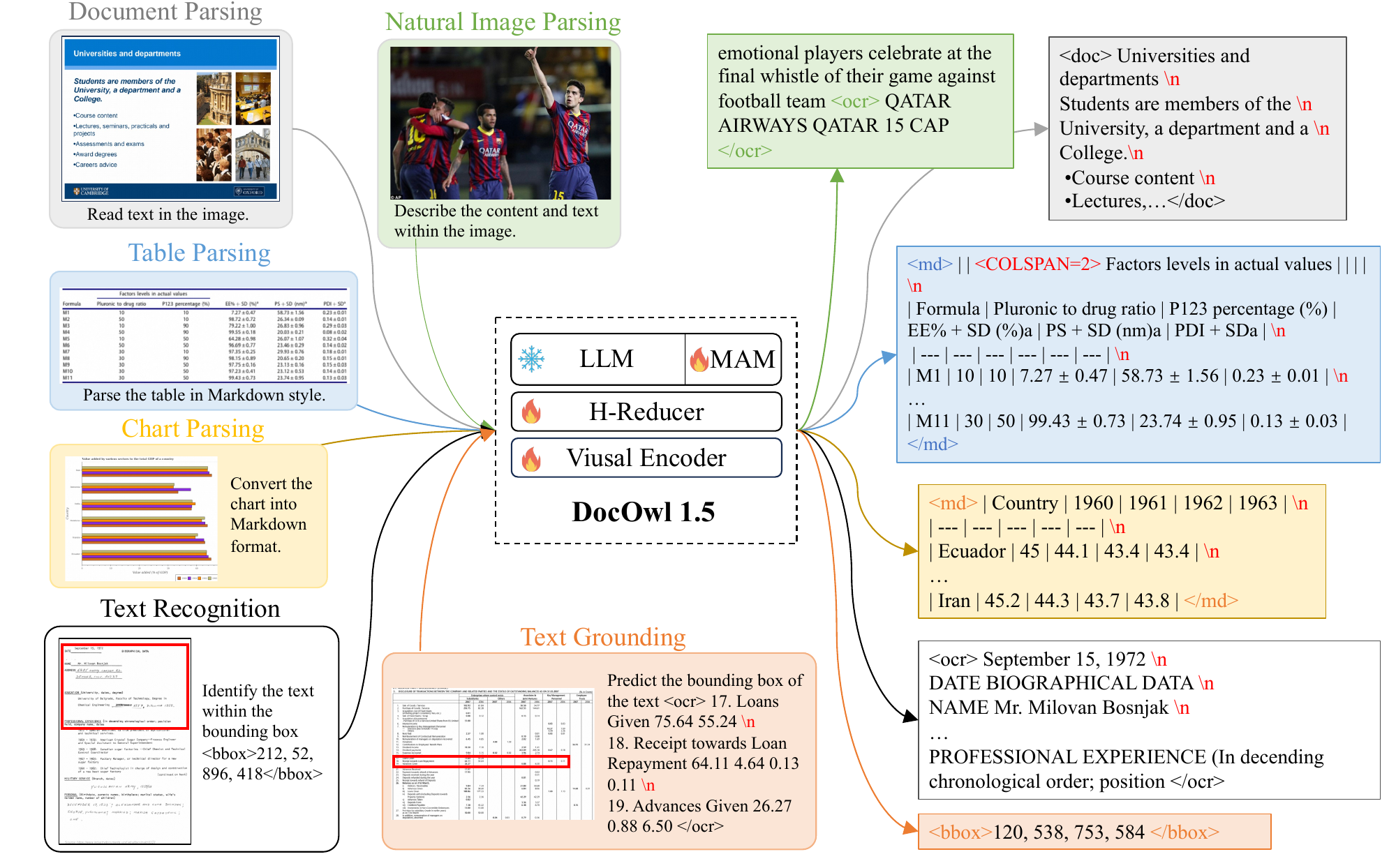}
    \caption{The illustration of Unified Structure Learning of \modelname.} 
    \label{fig:layout_tasks}
\end{figure*}

 Most Multimodal Large Language Models~\citep{llava,mplug-owl2,cogvlm} are trained with image-text pairs of natural images to align the visual encoder with the LLM, such as Conceptual Captions~\citep{ConceptualCaption}, LAION~\citep{laion} and COYO~\citep{coyo}. Initializing from such models could inherit the shallow text recognition ability, but is far from understanding complex textual and structural information in various text-rich images. In this work, to empower the comprehensive document understanding abilities of MLLM, we design a Unified Structure Learning across 5 domains, including natural images, documents, tables, charts, and webpages. It involves both structure-aware parsing tasks and multi-grained text localization tasks, as shown in \cref{fig:layout_tasks}.

\noindent\textbf{Document Parsing.} 
For representing the structure information, Pix2Struct~\citep{pix2struct} parses webpage screenshots with condensed HTML DOM trees, which are built based on the HTML source codes and are not available for other formats of documents or webpage screenshots, e.g. PDF. In documents or webpages, horizontal and vertical distances between texts form the main layout information. Therefore, to make the structure-aware parsing task applicable to most documents and webpage screenshots, we choose to add extra line feeds(\texttt{`$\textbackslash n$'})  and spaces into the text sequence to denote different lines and horizontal distances. The greater the horizontal distance, the more space characters. 

We choose CCpdf~\citep{ccpdf}, RVL-CDIP~\citep{rvlcdip},  VisualMRC~\citep{visualmrc} and datasets encapsulated in DUE-Benchmark~\citep{due} (DocVQA~\citep{docvqa}, InfoVQA~\citep{infovqa}, DeepForm~\citep{deepform}, KLC~\citep{klc}, WTQ~\citep{wikitableqa}, TabFact~\citep{TabFact}) to support the Document Parsing task. CCpdf~\citep{ccpdf} is a multi-lingual PDF dataset built upon webpages from Common Cramwl\footnote{\url{https://commoncrawl.org}}, covering diverse domains of documents, such as industry, academic, and medical. In this work, we mainly focus on English Document Understanding and drop PDFs detected as other languages. RVL-CDIP contains 16 categories of industry documents, such as `letter', `email', and `scientific reports'. We further remove some categories with flipping and blurring texts, such as `handwritten' and `form'. DUE-Benchmark is a collection of available and reformulated datasets over various document domains and layouts featuring tables, graphs, lists, and infographics. VisualMRC is a webpage screenshot dataset across 35 websites. OCR annotations in VisualMRC are aligned with local regions, thus, we follow them to utilize crops of a screenshot as input for this parsing task. 
For CCpdf and DUE-Benchmark, a PDF-parsing tool pdfplumber\footnote{\url{https://github.com/jsvine/pdfplumber}} can be directly used to generate structure-aware text sequence with a PDF page as the input. For RVL-CDIP and VisualMRC, there are no PDF files, just annotations of bounding boxes of texts. 
As an alternative, akin to the LATIN-Prompt~\citep{latin}, we insert the line feeds and spaces by calculating and comparing the horizontal and vertical distances of bounding boxes. To avoid too many space characters resulting in sparse texts, we further limit the maximum number of consecutive spaces to 4. This strategy allows us to construct structure-aware text sequences in the same style as pdfplumber.

\noindent\textbf{Table Parsing.} Different from documents or webpages, tables are structured in a more standardized way, where row and column correspondences represent key-value pairs. HTML and Markdown codes are mainly two kinds of text sequences used to represent a table. HTML codes can represent all kinds of tables, with or without cells spanning multiple rows and grids, but they contain too many paired labels (e.g. \texttt{`<tr></tr>'} and \texttt{`<td></td>'}), causing text sequences to be too long. Markdown codes can represent a table with concise text sequence, but they cannot represent cells spanning multiple rows and columns. To represent all tables with concise text sequence, we follow the main grammar of Markdown to represent table structure with \texttt{`|'} and line feeds(\texttt{`$\textbackslash n$'}). To represent cells spanning multiple rows and columns, we add special text tokens \texttt{`<COLSPAN=x>'} and \texttt{`<ROWSPAN=y>'} before the value, as shown in \cref{fig:layout_tasks}. 

We choose TURL~\citep{turl} and PubTabNet~\citep{pubtabnet} to do the structure-aware table parsing task, where tables are collected from Wikipedia pages and scientific articles, respectively. Without cells across rows and columns, tables in TURL can be directly represented with Markdown codes. Due to lacking table images in TURL, we transfer tables into HTML codes and render table images with variations in background color and font size. PubTabNet contains pairs of table images and HTML codes. We convert HTML codes into Markdown style and add \texttt{`<ROWSPAN=x>'} or \texttt{`<COLSPAN=y>'} before the value when attributes \texttt{`rowspan=x'} or \texttt{`colspan=y'} are set in the \texttt{`<td>'} label.

\noindent\textbf{Chart Parsing.} Unlike documents and tables, organizing texts in reading order cannot represent the structure of charts. Considering that the chart is a visualization form of the table, parsing charts to tables could best maintain the mathematical characteristics of the chart. This requires the model to understand the structure of the chart and the alignment of the x/y axis. Besides, to keep consistent with the Table Parsing task, we also use Markdown codes to represent the data tables of charts, as shown in \cref{fig:layout_tasks}.

We adopt PlotQA~\citep{plotqa}, FigureQA~\citep{figureqa}, DVQA~\citep{dvqa}, and ChartQA~\citep{chartqa} to support the structure-aware chart parsing task. These datasets cover charts on both synthetic~\citep{figureqa, dvqa} data and data from real-world sources~\citep{plotqa,chartqa}. Chart types include vertical bar, horizontal bar, line, dot line, and pie chart. Source data of the chart is provided in the JSON~\citep{plotqa,figureqa,plotqa} or CSV format~\citep{chartqa}, both can be conveniently converted to Markdown codes. However, some raw values are not suitable as standard answers for parsing because there are too many significant digits to be represented on the chart. Therefore, to reduce the difficulty of estimating values and make the model focus more on structural understanding, we keep 4 significant digits for all values.

\noindent\textbf{Natural Image Parsing.} Quite different from text-dominant images mentioned above, the semantics of natural images is a combination of natural objects and scene texts. Thus, parsing natural images is necessary to organize scene texts and mention the main image content.  Manually annotating captions to describe the relationship between objects and scene texts is labour- and financial-intensive. Like TAP~\citep{tap}, we concatenate the general caption with OCR texts to form the target parsing sequence.

We utilize OCR-CC~\citep{tap} to support the Natural Image Parsing task. OCR-CC is a subset of Conceptual Caption~\citep{cc2018}, which contains images with scene texts detected by the Microsoft Azure OCR system.

\noindent\textbf{Multi-grained Text Localization.} As proved in previous works~\citep{e2evlp,ofa,kosmos2} on general image understanding, semantic comprehension and object grounding tasks can be well unified in a single model. For Visual Document Understanding, structure-aware parsing tasks mainly focus on organizing texts according to the overall structure, while neglecting the correspondence between specific texts and local positions. Correlating texts with the concrete position in images is another basic structure understanding ability for visual documents. To support text position learning, we design two symmetrical tasks, namely Multi-grained Text Grounding and Multi-grained Text Recognition. The former aims to predict the bounding box given the visually-situated texts, while the latter does the opposite. We set four granularities of texts for these two tasks: word, phrase, line, and block. The `word' is the smallest granularity of the bounding box, referring to only 1 word. To ensure that the word is visible and the answer is unique, words that are too small (normalized area < 0.001) and words that appear multiple times in the same image are excluded from candidates. The `line' consists of texts that are judged to be horizontally parallel by vertical distance, and the `phrase' is comprised of multiple adjacent words within the same line. The `block' is a combination of multiple successive lines, ranging from 2 to half of the total lines. The text sequences of word-level and phrase-level question answering are much shorter than the other two. Therefore, in order to learn localization more efficiently, each word-level or phrase-level sample consists of up to 5 question-answer pairs for the same image. As for the representation of bounding boxes, we transfer each continuous value in the normalized bounding box into a discrete position token, ranging from 0 to 999. 

The bounding box annotation is necessary for constructing samples for Multi-grained Text Localization tasks. Therefore, we take DocVQA, InfoVQA, WTQ, TabFact, DeepForm, KLC, ChartQA, VisualMRC, and TextVQA~\citep{textvqa} for this task, across domains of the document, table, chart, webpage, and natural image. 

\begin{figure*}[tp]
    \centering
    \includegraphics[width=0.9\linewidth]{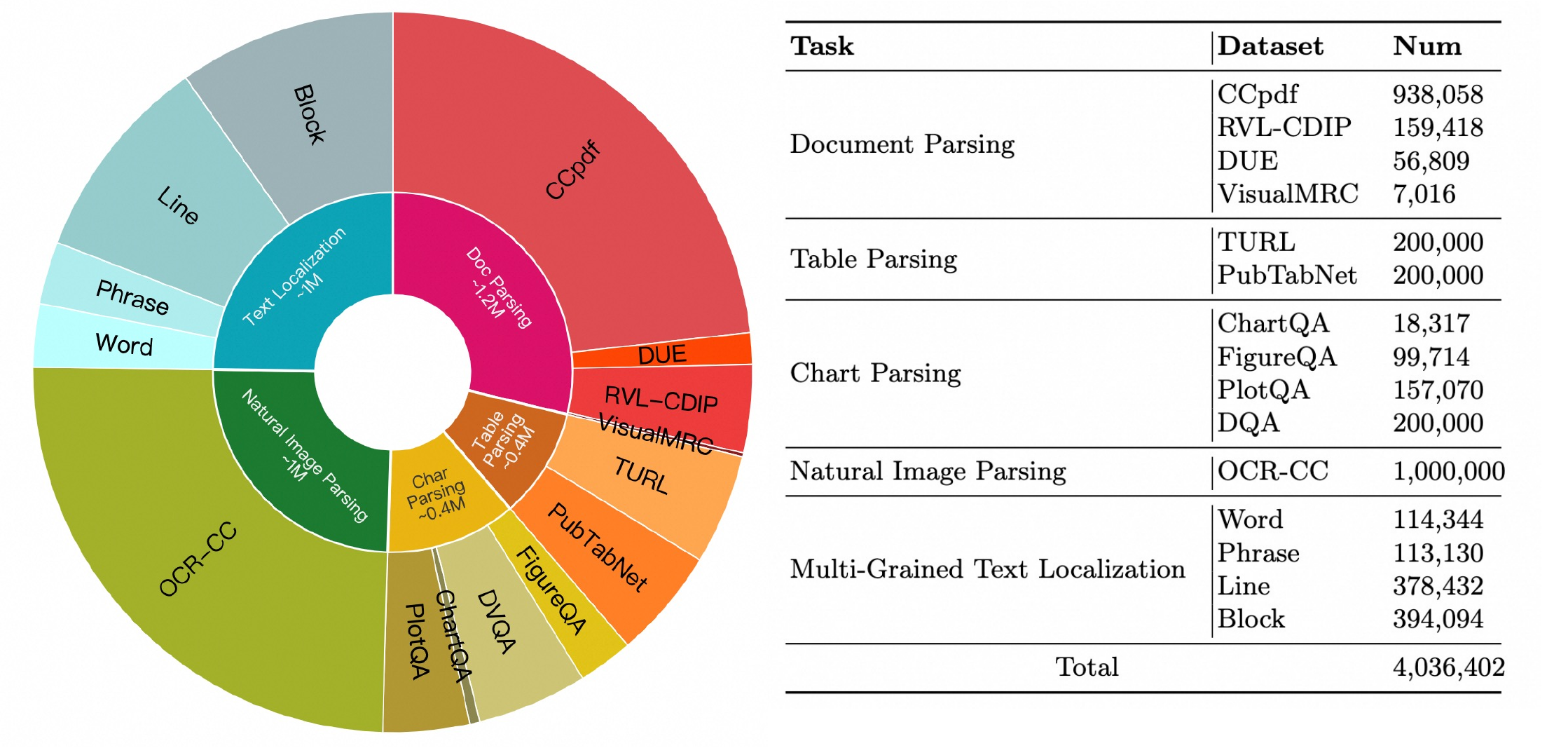}
    \caption{Detailed statistics of \pretraindataname.} 
    \label{fig:data_distri}
\end{figure*}

Overall, to support the unified structure learning for text-rich images, we build a \pretraindataname~dataset by ensembling multiple training sets of publicly available datasets and constructing structure-aware text sequences or text-position pairs as the targets. The form of instructions for each task is very diverse for developing the general instruction-following ability of the model.
\cref{fig:data_distri} shows the detailed statistics of \pretraindataname.

\subsection{Multi-task Fine-tuning}
 Through Unified Structure Learning, models could well understand the structure of diverse document images but cannot follow users' instructions to do different types of tasks, such as information extraction or image captioning. So, we further perform multi-task fine-tuning to train a generalist of visual document understanding as UReader~\citep{ureader}.

 \subsection{Training Paradigm}
 As shown in \cref{fig:overall_arch}(a), \modelname~is trained in a two-stage framework.
 Considering the LLM has strong  comprehension abilities for structured text~\citep{latin, tablellama}, we argue that the main limitation of MLLM in visual document understanding is the representation ability of the Visual Encoder and Vision-to-Text module for visually-situated text and structure information. Thus, during the Unified Structure Learning, we freeze the LLM parameters and tune the Visual Encoder and \connectname. The MAM is also optimized to help the LLM better distinguish visual features and texts parsed from the image. During the stage of Multi-task Fine-tuning, the model mainly learns how to follow the user's instructions to give answers based on visually-situated text and structure understanding capabilities acquired in the first stage. Therefore, the Visual Encoder is frozen and other modules are tuned.

\section{\modelname-Chat}
Existing benchmarks mainly evaluate the document understanding ability by answering the question with simple phrases and neglect detailed explanations. In this work, to better leverage the strong language reasoning ability of Large Language Models on Visual Document Understanding, we build a small instruction-tuning set with detailed explanations on text-rich image understanding, namely \instructset. Based on raw questions from DocVQA~\citep{docvqa}, InfoVQA~\citep{infovqa}, WTQ~\citep{wikitableqa}, VisualMRC~\citep{visualmrc}, ChartQA~\citep{chartqa} and TextVQA~\citep{textvqa}, we collect detailed explanations with ChatGPT\footnote{\url{https://openai.com/chatgpt}}. Text contents are dominant information on documents, tables or webpage screenshots. Therefore, for DocVQA, InfoVQA, WTQ, and VisualMRC, we take the structure-aware text sequence of the image as the input to \texttt{gpt-3.5-turbo-0301} and prompt it to answer the question with simple answers and detailed explanations. As for ChartQA and TextVQA, we take the image as the input and utilize the \texttt{gpt-4-vision-preview} to answer the question with detailed explanations. In order to filter out samples where ChartGPT answers incorrectly, we further prompt \texttt{gpt-3.5-turbo-0301} to judge whether the answer given by ChartGPT is consistent with the concise human-annotated ground-truth answer. Compared with raw questions in benchmark datasets, questions in \instructset~are added with a prompt \texttt{`Answer the question with detailed explanation'}. Detailed statistics~of \instructset~are presented in \cref{tab:instruct_set}. \modelname-Chat is trained by combining downstream datasets with \instructset~and performing multi-task tuning after Unified Structure Learning.

\begin{table*}
    \caption{The detailed statistics of \instructset. The `Avg Length' refers to the average token length of the answer.}
    \label{tab:instruct_set}
    \footnotesize
    \centering
    \begin{tabular}{c|cccccc|c}
    \toprule
    ~ & DocVQA & InfoVQA & WTQ & VisualMRC & ChartQA & TextVQA & ALL\\
    \midrule
    Image & 1,491 & 1,614 & 850 & 1,927 & 1,252 & 1,612 & 8,746 \\
    Sample & 5,119 & 5,421 & 5,994 & 5,263 & 1,827 & 2,253 & 25,877 \\
    Avg Length & 79.2 & 95.4 & 77.7 & 103.4 & 106.9 & 88.0 & 89.9 \\
    \bottomrule
    \end{tabular}
\end{table*}

\section{Experiments}

\subsection{Implementation Details}
\modelname~is initialized from mPLUG-Owl2~\citep{mplug-owl2}, which utilizes the ViT/L-14~\citep{vit2021} as the Visual Encoder and a 7B Large Langauge Model with the Modality Adaptive Module as the language decoder. According to the aspect ratio and resolution, each image is cropped into up to 9 sub-images with a fixed resolution of 448x448. Each sub-image is encoded to 1,024 features by the ViT/L-14 and then reduced to 256 features by the \connectname. The model is trained with 12,000 iterations on \pretraindataname, with the learning rate and batch size set as 1e-4 and 1,024. It costs about 128 A100 days. During the Multi-task finetuning, the model is trained for 6,500 iterations with the batch size set as 256 and the learning rate set as 2e-5. This further costs about 24 A100 days.

\begin{table*}[t]
    \caption{Different settings of OCR-free Visual Document Understanding models. `Open' refers to whether all OCR learning data is open-source.}
    \label{tab:model_setting}
    \footnotesize
    \centering
    \resizebox{\linewidth}{!}{
    \begin{tabular}{l|lc|ccccc}
    \toprule
    \multirow{2}*{\textbf{Model}} & \multirow{2}*{\textbf{Init}} & \multirow{2}*{\textbf{Resolution}} & \multicolumn{5}{c}{\textbf{OCR Learning}} \\
    ~ & ~ & ~ & \textbf{Text} & \textbf{Bbox} & \textbf{Size}  &\textbf{Domain} &\textbf{Open} \\
    \midrule
    Donut~\citep{donut} & - & 2560x1920 & $\checkmark$ & $\times$ & 13M & Synthetic, Doc & $\checkmark$ \\
    Pix2Struct~\citep{pix2struct} & - & $2^{19}$(shape variable) & $\checkmark$ & $\times$ & 80M  & Web & $\times$ \\
    QwenVL~\citep{qwenvl} & - & 448x448 & $\checkmark$ & $\times$ & 24.8M & Synthetic, Doc, Web & $\times$ \\
    Monkey~\citep{monkey} & QwenVL~\citep{qwenvl} & 896x896 & $\times$ & $\times$ & - & - & - \\
    UReader~\citep{ureader} & Owl~\citep{mplugowl} & 224x224(x20 crops) & $\checkmark$ & $\times$ & 0.1M  & Doc, Table, Chart, Web, Natural & $\checkmark$ \\
    DocPedia~\citep{docpedia} & - & 2560×2560 & $\checkmark$ & $\checkmark$ & 0.9M  & Doc & $\times$ \\
    CogAgent~\citep{cogagent} & CogVLM~\citep{cogvlm} & 1120×1120 & $\checkmark$ & $\checkmark$ & 107M  & Synthetic, Nature, Doc, Web & $\times$ \\
    \midrule
    DocOwl 1.5 & Owl2~\citep{mplug-owl2} & 448x448(x9 crops) & $\checkmark$ & $\checkmark$ & 4M  & Doc, Table, Chart, Web, Natural & $\checkmark$ \\
    \bottomrule
    \end{tabular}
    }
\end{table*}

\begin{table*}
    \caption{Comparison with OCR-free methods on various types of text-rich image understanding tasks. The superscript `$*$' refers to models separately fine-tuned on each downstream task, rather than generalists. The $\underline{underline}$ means the best performance among models with <10B parameters.}
    \label{tab:main}
    \footnotesize
    \centering
    \resizebox{\linewidth}{!}{
    \begin{tabular}{lc|cccc|cc|c|cc|c}
    \toprule
    \multirow{2}*{\textbf{Model}} & \multirow{2}*{\textbf{Size}} & \textbf{Doc} & \textbf{Info} & \textbf{Deep} & \multirow{2}*{\textbf{KLC}} & \multirow{2}*{\textbf{WTQ}}  & \textbf{Tab} & \textbf{Chart} & \textbf{Text} & \textbf{Text} & \textbf{Visual} \\ 
    ~ & ~  & \textbf{VQA} & \textbf{VQA} & \textbf{Form} & ~ & ~ & \textbf{Fact} & \textbf{QA} & \textbf{VQA} & \textbf{Caps} & \textbf{MRC} \\
    \midrule
    Dessurt$^{*}$ & <1B & 63.2 & -& - & - & - & - & - & - & - & - \\ 
    Donut$^{*}$ & <1B &67.5 & 11.6 & 61.6 & 30.0 & 18.8 & 54.6 &41.8 & 43.5 & 74.4 & 93.91 \\
    Pix2Struct$_{base}^{*}$ & <1B & 72.1 & 38.2 &- & - & - & - & 56.0 & -& 88.0 & -  \\ 
    Pix2Struct$_{large}^{*}$ & 1.3B & 76.6 & 40.0 & - & - & - & - & 58.6 & -& 95.5 & -  \\ 
    \midrule
    DocPeida & 7.0B & 47.1 & 15.2 & - & - & - & - & 46.9 & 60.2 & - & - \\
    DocOwl & 7.1B & 62.2 & 38.2 & 42.6 & 30.3 & 26.9 & 60.2 & 57.4 & 52.6 & 111.9 & 188.8 \\
    QwenVL &9.6B & 65.1 & 35.4 & - & - & - &-& 65.7 & 63.8 & - & -  \\
    UReader &7.1B & 65.4 & 42.2 & 49.5 & 32.8 & 29.4  & 67.6 & 59.3 & 57.6 & 118.4 &221.7 \\
    Monkey & 9.8B & 66.5 & 36.1 & 40.6 & 32.8 & 25.3 & - & - & 67.6 & 93.2 & - \\
    CogAgent & 17.3B & 81.6 & 44.5 & -& -&- & - & 68.4 & \textbf{76.1} & - & -  \\
    \midrule
    DocOwl-1.5 &8.1B  & 81.6 & 50.4 & 68.8 & 37.9 & 39.8  & \textbf{80.4} & \textbf{70.5} & \underline{68.8} & \textbf{132.0} & 239.5 \\
    DocOwl-1.5-Chat &8.1B  & \textbf{82.2} & \textbf{50.7} & \textbf{68.8} & \textbf{38.7} & \textbf{40.6}  & 80.2 & 70.2 & 68.6 & 131.6 & \textbf{246.4} \\
    \bottomrule
    \end{tabular}
    }
\end{table*}

\subsection{Main Results}
We evaluate the Visual Document Understanding performance on 10 text-rich image benchmarks, covering documents (DocVQA~\citep{docvqa}, InfoVQA~\citep{infovqa}, DeepForm~\citep{deepform}, KLC~\citep{klc}), tables (WTQ~\citep{wikitableqa}, TabFact~\citep{TabFact}), charts (ChartQA~\citep{chartqa}), natural images (TextVQA~\citep{textvqa}, TextCaps~\citep{textcaps}), and webpage screenshots (VisualMRC~\citep{visualmrc}). We compare \modelname~with state-of-the-art OCR-free models, including both Multimodal Large Language Models adapted for recognizing texts and much smaller models trained only for document understanding. The detailed comparison of model settings can be found in \cref{tab:model_setting}. As shown in \cref{tab:main}, previous MLLMs with more than 7B parameters underperform domain-specific models with less than 1B parameters, showing that the document understanding is still a shortcoming for existing MLLMs. Our \modelname~outperforms both domain-specific models and MLLMs with similar sizes on all 10 benchmarks. This validates that \modelname~is much stronger on visual document understanding across 5 domains, covering visual question answering, information retrieval, natural language inference, and image captioning tasks. Besides, with much fewer unnatural data (3M vs 9M) and parameters (8.1B vs 17.3B), \modelname~outperforms CogAgent~\citep{cogagent} on InfoVQA and ChartQA, and achieves comparable performance on DocVQA. This suggests that our unified structure learning with \pretraindataname~is more efficient in learning printed text recognition and how to analyze documents. However, our model still underperforms CogAgent on TextVQA, which requires the ability of scene text recognition and general knowledge about natural objects. 
The primary reason is that scene texts are more diverse in shapes than printed texts and CogAgent is trained on 98M samples of scene text recognition from LAION-2B~\citep{laion} and COYO-700M~\citep{coyo}, much more than the natural images (1M) in \pretraindataname. 
In this work, we mainly focus on improving the unified structure comprehension of visual documents and leave further scaling up data on natural scenes as future work. Finally, \modelname-Chat can also be evaluated on these concise-answer benchmarks by removing the prompt of detailed explanation. It achieves comparable or slightly better performance than \modelname, showing that a small amount of detailed explanatory data may better help the model understand the semantics of text-rich images.

\begin{table*}
    \caption{Ablation study of model setting. `Crop' refers to the maximum number of cropped images. `CropPos' means using learnable embeddings (`Emb') or textual tokens (`Text') to represent the position of cropped images. `Parsing' and `MTL' refer to structure-aware parsing tasks and the Multi-grained Text Location task, respectively. `Owl(224)' and `Owl2(448)' refer to mPLUG-Owl~\citep{mplugowl} with 224 resolution and mPLUG-Owl2~\citep{mplug-owl2} with 448 resolution, respectively.}
    \label{tab:ablation}
    \footnotesize
    \centering
    \resizebox{\linewidth}{!}{
    \begin{tabular}{c|cccc|c|cc|ccc}
    \toprule
    ~ & \multicolumn{4}{c|}{\textbf{Model Architecture}} & \textbf{Structure} & \multicolumn{2}{c|}{\textbf{Multi-task Tuning}} & \multirow{2}*{\textbf{DocVQA}}  & \multirow{2}*{\textbf{TabFact}} & \multirow{2}*{\textbf{ChartQA}}  \\
     ~ & \textbf{Init} & \textbf{V2T} & \textbf{Crop} & \textbf{CropPos} & \textbf{Learning} &\textbf{ViT} & \textbf{LLM} & ~ & ~ & ~  \\
    \midrule
    r1 & Owl(224) & Abstractor & 20 & Emb & $\times$ & $\times$ & $\times$  & 65.4  & 67.6 & 59.3 \\ 
    r2 & Owl2(448) & Abstractor & 20 & Emb & $\times$ & $\times$ & $\times$ & 66.3 & 69.8 & 60.6 \\
    r3 & Owl2(448) & Abstractor & 20 & Emb & $\times$ & $\checkmark$  & $\times$ & \cellcolor{Gray}71.4 & \cellcolor{Gray}70.3  & \cellcolor{Gray}64.2 \\ 
    r4 & Owl2(448) & Abstractor & 9 & Emb  & $\times$ & $\checkmark$ & $\times$ &  68.0 & 70.0 & 64.2 \\ 
    \midrule
    r5 & Owl2(448) & \connectname(1x4) & 9 & Emb  & $\times$ & $\checkmark$ & $\times$ & \cellcolor{Gray}72.8 & \cellcolor{Gray}72.9 & \cellcolor{Gray}65.0 \\
    r6 & Owl2(448) & \connectname(2x2) & 9 & Emb  & $\times$ & $\checkmark$ & $\times$ & 71.8  & 72.1 & 65.2 \\
    r7 & Owl2(448) & \connectname(2x4) & 9 & Emb  & $\times$ & $\checkmark$ & $\times$ & 71.4 & 71.1 & 66.0 \\
    r8 & Owl2(448) & \connectname(1x8) & 9 & Emb  & $\times$ & $\checkmark$ & $\times$ & 69.9 & 71.2 & 64.4 \\
    r9 & Owl2(448) & \connectname(2x8) & 9 & Emb  & $\times$ & $\checkmark$ & $\times$ & 69.2 & 70.2 & 65.6 \\
    \midrule
    r10 & Owl2(448) & \connectname(1x4) & 9 & Emb & Parsing & $\times$ & $\times$ & 77.7 & 76.5 & 67.5 \\
    r11 & Owl2(448) & \connectname(1x4) & 9 & Emb  & Parsing & $\times$ & $\checkmark$ & 78.9  & 78.1 & 68.1 \\
    r12 & Owl2(448) & \connectname(1x4) & 9 & Text & Parsing & $\times$ & $\checkmark$  & 79.8  & 77.7 & 69.1 \\
    r13 & Owl2(448) & \connectname(1x4) & 9 & Text  & Parsing+MTL & $\times$ & $\checkmark$ & \cellcolor{Gray}81.6  & \cellcolor{Gray}80.4 & \cellcolor{Gray}70.5 \\
    \bottomrule
    \end{tabular}
    }
\end{table*}

\vspace{-10pt}
\subsection{Ablation Study}
As shown in \cref{tab:ablation}, we further perform a comprehensive ablation study to validate the effectiveness of our \connectname~and Unified Structure Learning.

Firstly, initializing from a stronger general MLLMs brings better performance on text-rich images (r2 vs r1), showing general vision-and-language knowledge benefits visual document understanding. Tuning the visual encoder during multi-task fine-tuning significantly improves the document understanding performance (r3 vs r2). This suggests that the visual representation of document images may be the main shortcoming of MLLMs and inspires us to design Unified Structure Learning to enhance the representation ability of the visual encoder for visually situated texts and structure. 

\noindent\textbf{Effectiveness of \connectname.} When using the Shape-adaptive Cropping Module, the image resolution supported by the MLLM is the product of the cropping number and basic resolution of each crop. With the Abstractor as the vision-to-text module, reducing the cropping number causes an obvious performance decrease (r4 vs r3) on documents. However, with a smaller cropping number, the \connectname~achieves better performance than the Abstractor (r5 vs r3), showing that $448^2\times9\approx2^{21}$ is an acceptable resolution for existing benchmarks and the \connectname~is stronger on maintaining rich text information during vision-and-language feature alignment. Besides, we further compare different settings of the merging shape in the convolution layer. With the same number of merged tokens, the model with the 1x4 merging shape achieves better performance than the one with the 2x2 merging shape on document and table datasets but slightly worse performance on chart understanding (r6 vs r5). This is consistent with the common sense that documents and tables mainly organize texts in the left-to-right order while the semantic structures of charts are much more flexible. A square merging shape is more suited to encode visual features in the form of bars, lines, or pies while the 1x4 merging shape is more appropriate for general document understanding. As shown in r7-r9, further extending the 1x4 merging shape horizontally and vertically decreases the length of visual features but at the cost of performance degradation. Considering the overall performance on all text-rich images, we finally choose the 1x4 as the merging shape in \connectname. 

\noindent\textbf{Effectiveness of Unified Structure Learning.} After determining the vision-to-text module, we perform two-stage training with Unified Structure Learning. With only the structure-aware parsing tasks, there is significant improvement across different domains (r10 vs r5). This validates that fine-tuning the visual encoder and \connectname~with structure-aware parsing tasks greatly helps MLLMs understand text-rich images. Further tuning the parameters of LLM brings slight improvement (r11 vs r10), suggesting that general language knowledge is not the main obstacle to visual document understanding. By replacing the learnable crop position embeddings with special textual tokens, the model achieves better performance (r12 vs r11), showing that the LLM can well understand the relative positions of multiple cropped images with just simple textual indicators. Finally, by introducing Multi-grained Text Localization tasks, \modelname~achieves the best performance, validating that correlating visually situated texts with concrete positions helps comprehend documents more accurately.

\begin{table*}[t]
    \caption{The comparison of two-stage training and one-stage joint training with increasing samples from \pretraindataname. For a fair comparison, the LLM is frozen for both two-stage and one-stage training. The bath size of one-stage training is always set as 256, the same as the Multi-task Tuning in two-stage training. }
    \label{tab:two_stage}
    \footnotesize
    \centering
    \begin{tabular}{l|ccccc|c}
    \toprule
    ~ & \multicolumn{5}{c|}{\textbf{One-Stage}} & {\textbf{Two-Stage}} \\
    \midrule
    \pretraindataname~samples& 0.0M & 0.5M & 1.0M & 2.0M & 4.0M & 4.0M \\
    Benchmark samples & 0.6M & 0.6M & 0.6M & 0.6M & 0.6M & 0.6M \\
    Epoch/iteration & 7/18k & 6/25k & 6/37k & 4/40k & 3/54k & 3/12k + 3/6.5k \\
    Cost (A100 days) & 60.0 & 83.3 & 123.3 & 133.3 & \cellcolor{Gray}180.0 & \cellcolor{Gray}144.8 \\
    \midrule
    DocVQA & 72.8 & 75.5 & 78.6 & 78.8 & \cellcolor{Gray}78.9 & \cellcolor{Gray}79.9 \\
    \bottomrule
    \end{tabular}
\end{table*}

\noindent\textbf{Effectiveness of the Two-stage Training.} As shown in \cref{tab:two_stage}, instead of two-stage training, we also try one-stage joint training of the structure learning and downstream tasks and gradually increase the samples from \pretraindataname. The epoch is gradually reduced because we didn't observe performance improvements with more iterations. For joint training, the model improves significantly on DocVQA as the samples of Unified Structure Learning increase when it is below 1M. However, as the Unified Structure Learning samples are further increased, the improvement of the model becomes subtle and its performance is not as good as the one using two-stage training. This shows that the two-stage training could better enhance basic text recognition and structure parsing abilities and is more beneficial and efficient for downstream document understanding. 

\begin{table*}
    \caption{The detailed statistic of \evalset.}
    \label{tab:eval_set}
    \footnotesize
    \centering
    \begin{tabular}{c|cccc|ccccc}
    \toprule
    \multirow{2}*{\textbf{Task}} & \multicolumn{4}{c|}{\textbf{Text Granularity}} & \multicolumn{5}{c}{\textbf{Image Domain}} \\
    ~ & \textbf{Word} & \textbf{Phrase} & \textbf{Line} &\textbf{Block} & \textbf{Doc} & \textbf{Table} & \textbf{Chart} &\textbf{Web} & \textbf{Natural}\\
    \midrule
    Text Recognition & 622 & 499 & 522 & 482 & 1,004 & 491 & 229 & 267 & 134 \\
    Text Grounding & 595 & 542 & 503 & 485 & 1,011 & 524 & 240 & 242 & 108 \\
    \bottomrule
    \end{tabular}
\end{table*}

\begin{table*}
    \caption{Multi-grained text localization performance of models with different vision-to-text modules.}
    \label{tab:grounding}
    \footnotesize
    \centering
    \resizebox{\linewidth}{!}{
    \begin{tabular}{cc|ccccc|ccccc}
    \toprule
    \multirow{2}*{\textbf{Module}} & \multirow{2}*{\textbf{Iter}} & \multicolumn{5}{c|}{\textbf{Text Grounding}} & \multicolumn{5}{c}{\textbf{Text Recognition}} \\
    ~ & ~ & \textbf{Word} & \textbf{Phrase} & \textbf{Line} &\textbf{Block} & \textbf{ALL} & \textbf{Word} & \textbf{Phrase} & \textbf{Line} &\textbf{Block} & \textbf{ALL}\\
    \midrule
    Abstractor & 1,800 & 10.92 & 25.83 & 34.59 & 87.01 & 37.69 & 30.68 & 28.58 & 40.12 & 32.73 & 33.03 \\
    \connectname(2x2) & 1,800 & 14.19 & 34.87 & 43.94 & 89.07 & 43.94 & 37.20 & 38.33 & 48.68 & 41.99 & 41.55 \\
    \connectname(1x4) & 1,800 & \textbf{17.82} & \textbf{39.30} & \textbf{53.28} & \textbf{90.52} & \textbf{48.28} & \textbf{39.60} & \textbf{41.84} & \textbf{55.37} & \textbf{49.84} & \textbf{46.66} \\
    \midrule
    \connectname(1x4) & 12,000 & 70.42 & 76.38 & 85.88 & 91.34 & 80.38 & 70.10 & 67.86 & 73.88 & 70.70 & 70.63 \\
    \bottomrule
    \end{tabular}
    }
\end{table*}

\subsection{Text Localization Evaluation}
Besides proving the effectiveness of \connectname~through downstream text-rich image understanding performance in \cref{tab:ablation}, we further directly compare the text localization performance after the Unified Structure Learning to validate its superiority in preserving spatial features. We build a text localization evaluation set \evalset~with 4,250 samples balanced on 4 granularities and covering both text recognition and text grounding tasks. The detailed statistics of \evalset~are shown in \cref{tab:eval_set}. Considering that document images are much more diverse and complex than other images, there are more samples in this domain than others. The IOU@0.5 is used to evaluate the text grounding performance. As for text recognition, the word, phrase, line, and block granularity is evaluated with BLEU1, BLEU2, BLEU3, and BLEU4~\citep{bleu}, respectively. As shown in \cref{tab:grounding}, when trained with the same iterations, the \connectname~achieves much better performance on both Text Recognition and Text Grounding tasks, showing that \connectname~with the 1x4 merging shape helps the LLM better understand concrete positions in images.

\begin{figure*}[tp]
    \centering
    \includegraphics[width=1.0\linewidth]{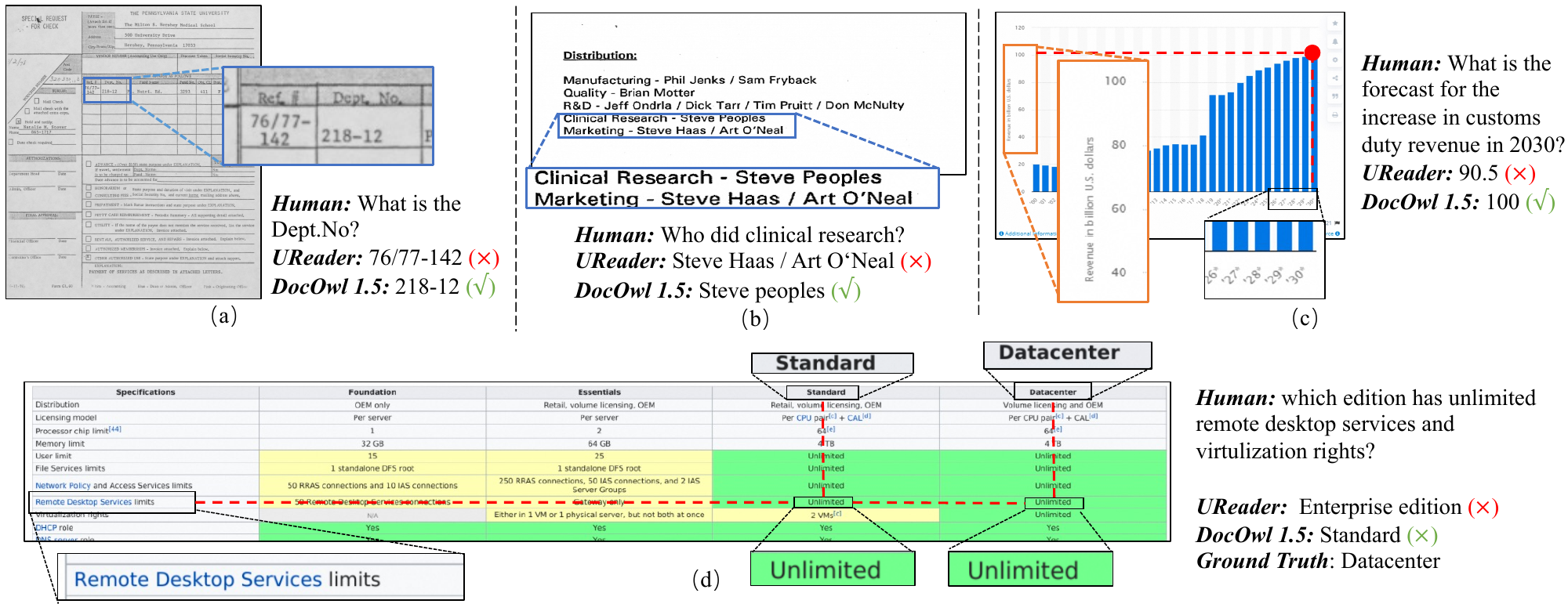}
    \caption{Qualitative results of \modelname~and UReader on different domains of images.} 
    \label{fig:qa_case}
\end{figure*}

\subsection{Qualitative Results}
\noindent{\textbf{Question Answering with Simple Phrases.}} Besides quantitative results, we further present some qualitative results of visual document understanding on different domains of images. As shown in \cref{fig:qa_case}(a) and (b), both models answer the question with texts in the image. \modelname~can better understand the structure of two documents and give correct answers. In \cref{fig:qa_case}(c), due to the learning of parsing chart with Markdown codes, \modelname~can better understand the chart and successfully correlate the x/y axis. \cref{fig:qa_case}(d) shows that although inconsistent with the ground truth, \modelname~gives another correct answer with the help of stronger structure understanding on tables.

\begin{figure*}[tp]
    \centering
    \includegraphics[width=1.0\linewidth]{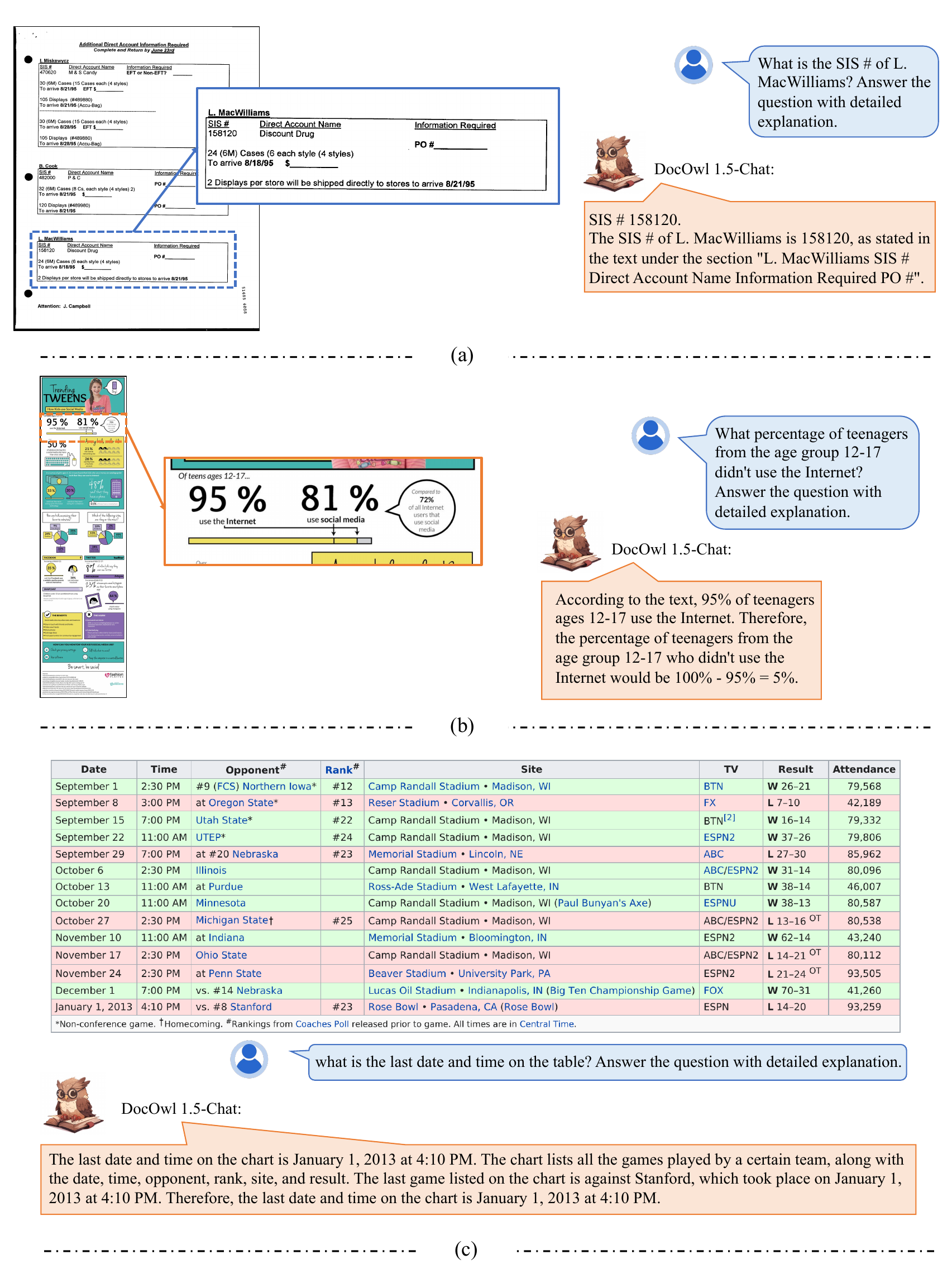}
    \caption{Qualitative results of question answering with detailed explanations. 
    Some regions are enlarged for better visualization.} 
    \label{fig:instruct_case_1}
\end{figure*}

\begin{figure*}[tp]
    \centering
    \includegraphics[width=1.0\linewidth]{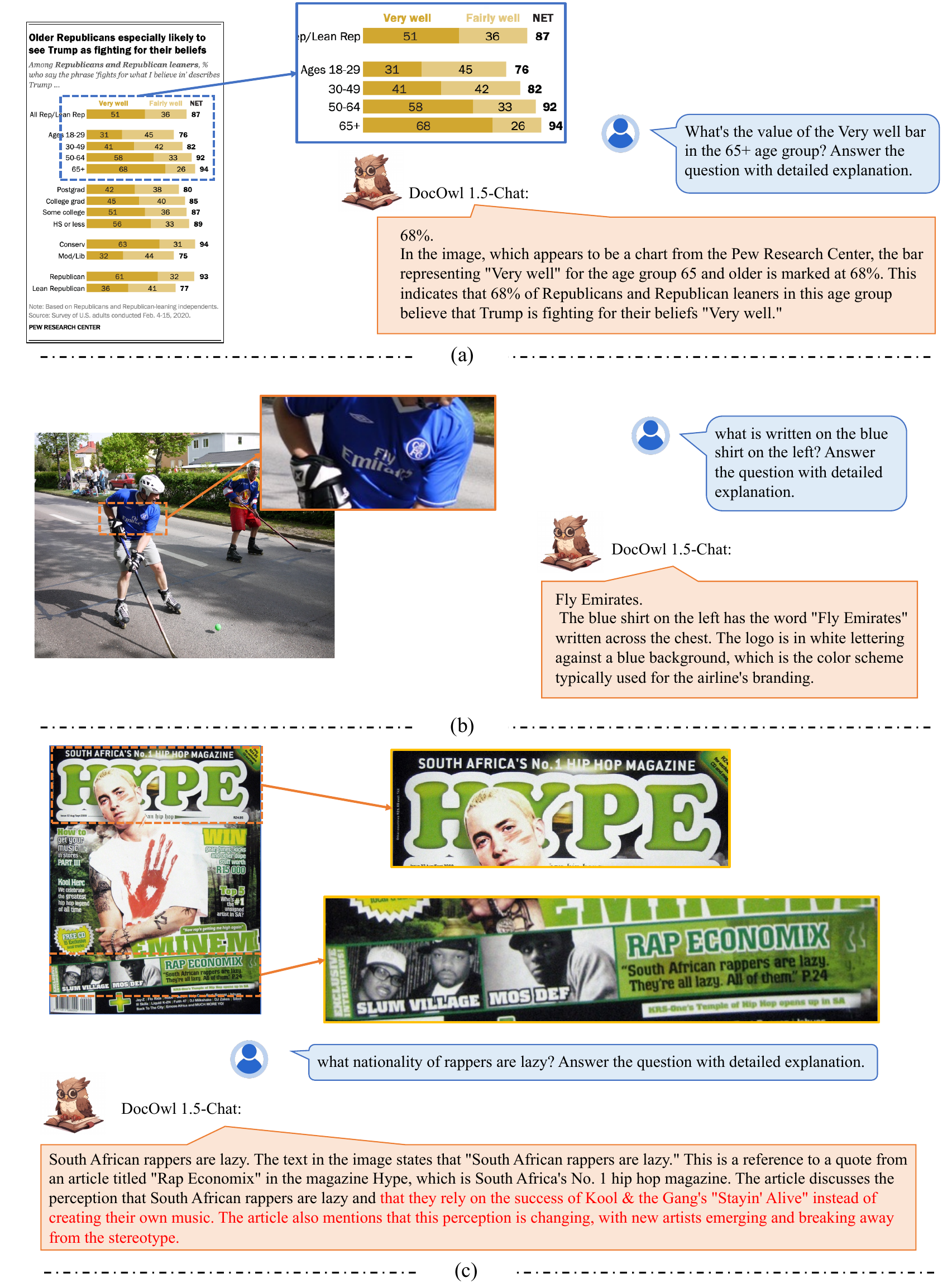}
    \caption{Qualitative results of question answering with detailed explanations. Hallucination in answers are marked in {\color{red}red}.} 
    \label{fig:instruct_case_2}
\end{figure*}

\noindent{\textbf{Question Answering with Detailed Explanations.}} \cref{fig:instruct_case_1} and \cref{fig:instruct_case_2} present qualitative results of detailed explanations. Through a small amount of reasoning training, \modelname-Chat can well inherit the reasoning ability of LLM and provide detailed explanations about the answer. However, as presented in \cref{fig:instruct_case_2}(c), like most general Multimoal large Language Models~\citep{mplugowl,mplug-owl2,qwenvl}, \modelname-Chat may also suffer from the hallucination problem in Visual Document Understanding. In this work, we mainly focus on enhancing the unified structure understanding ability of MLLMs and leave how to resolve the hallucination problem in OCR-free document understanding as future work.

\begin{figure*}[tp]
    \centering
    \includegraphics[width=1.0\linewidth]{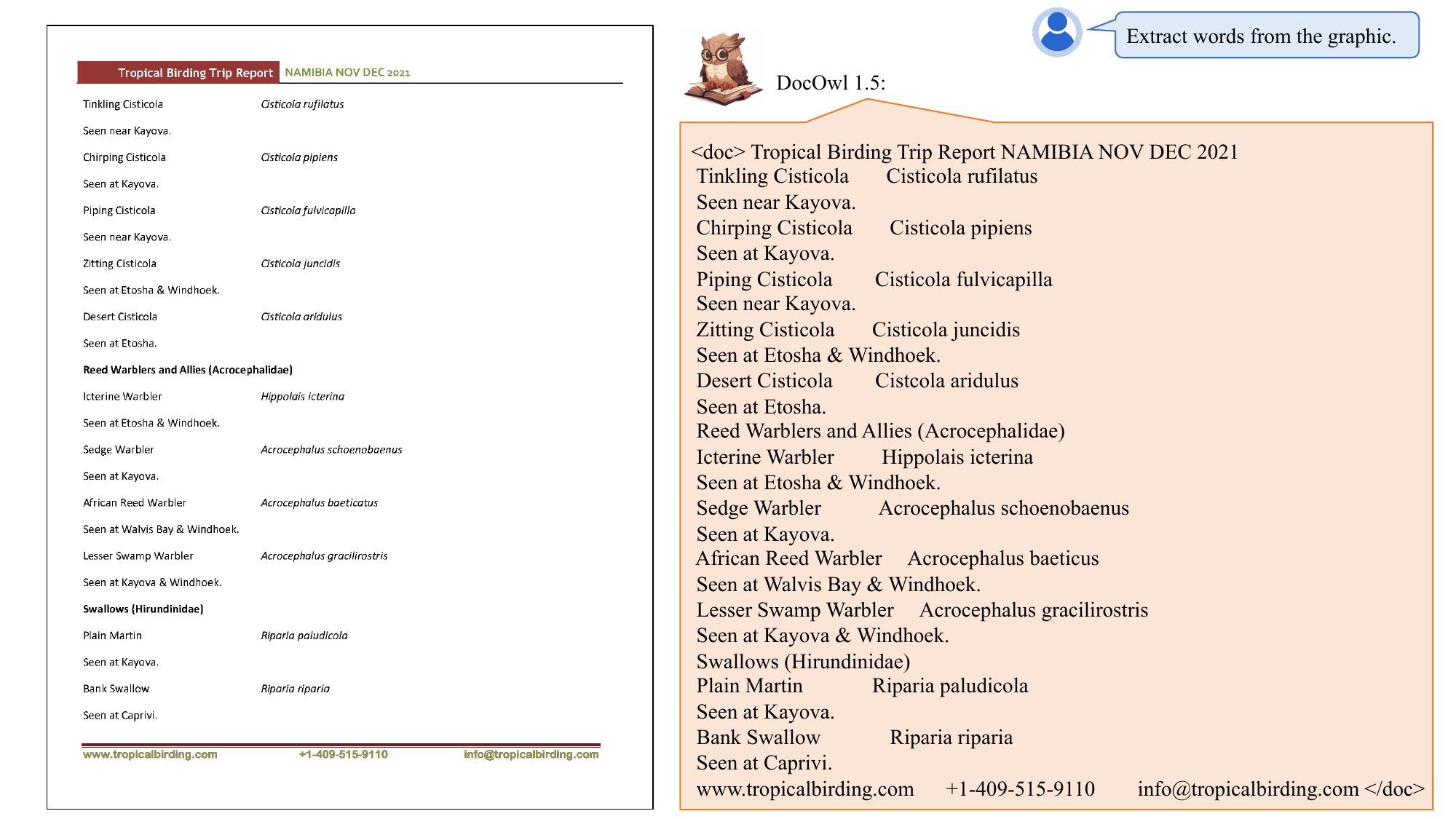}
    \caption{A qualitative result of structure-aware document parsing.} 
    \label{fig:doc_parse}
\end{figure*}

\begin{figure*}[tp]
    \centering
    \includegraphics[width=1.0\linewidth]{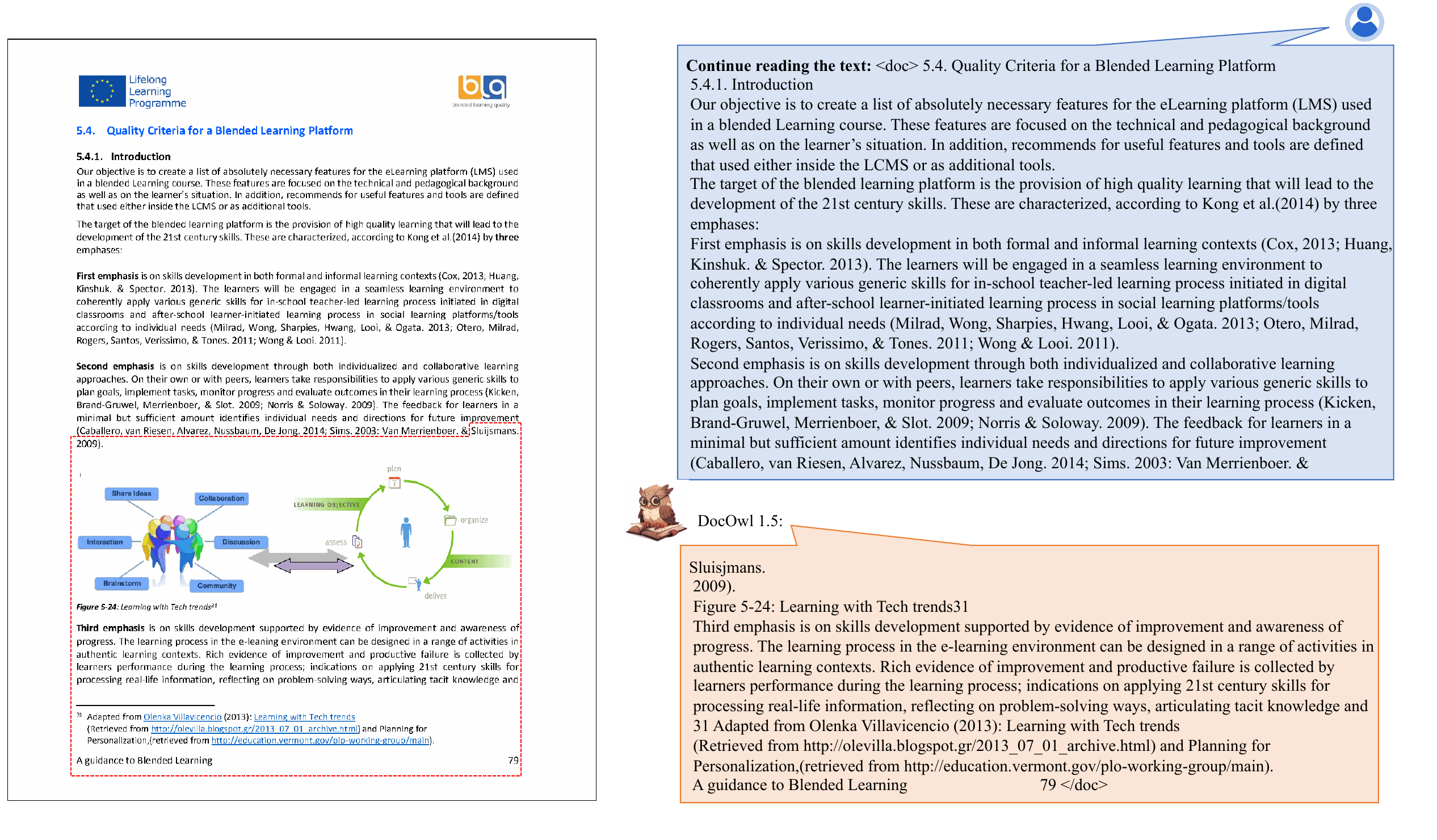}
    \caption{A qualitative result of structure-aware document parsing from the middle of the image. The red dotted box is only used to mark the location of the answers for better visualization and is not included in the input image.} 
    \label{fig:doc_parse2}
\end{figure*}

\begin{figure*}[tp]
    \centering
    \includegraphics[width=0.9\linewidth]{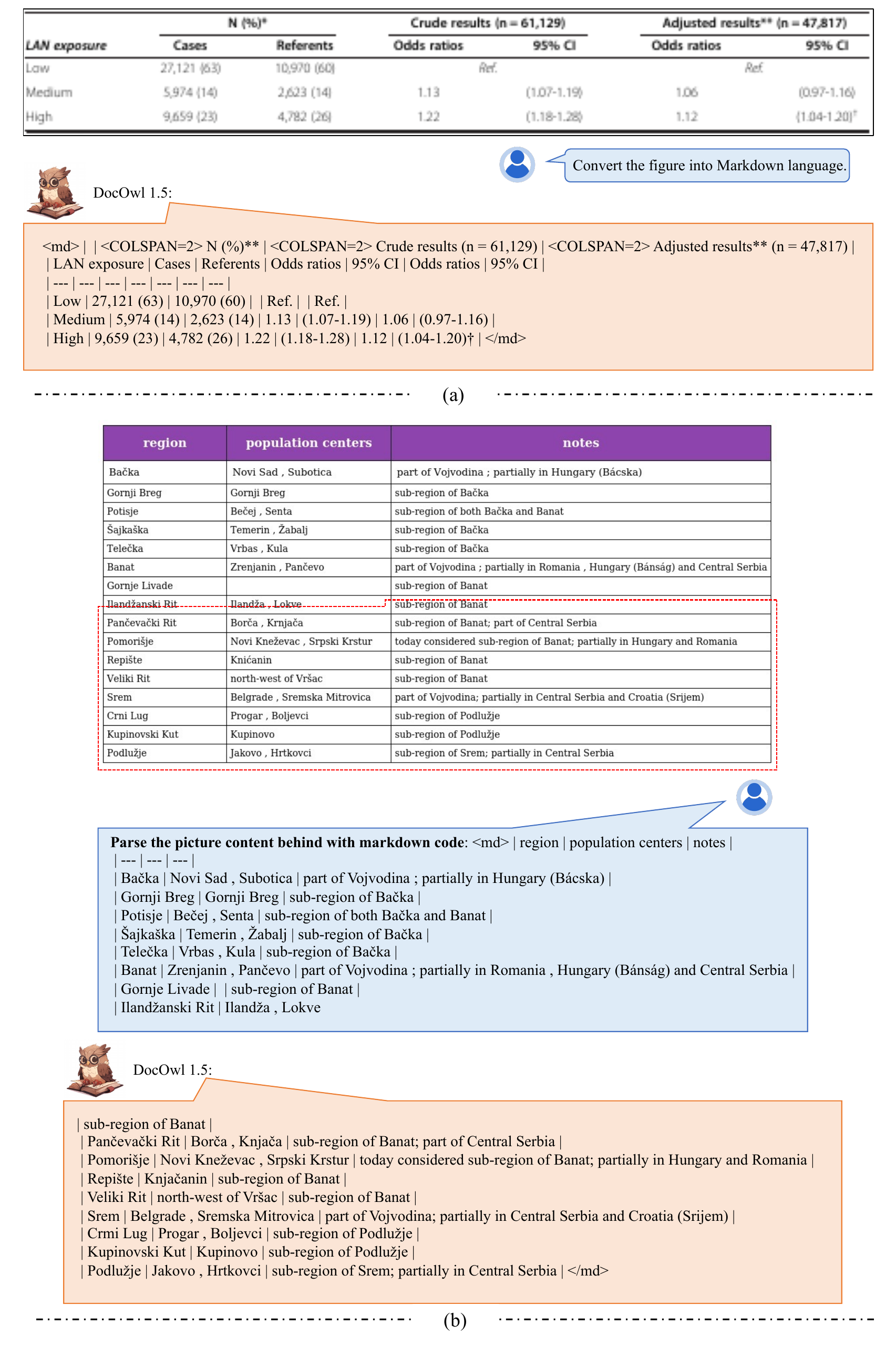}
    \caption{Qualitative results of structure-aware table parsing on the table with cells spanning multiple columns (a) and structure-aware table parsing from the middle of the image (b). The red dotted box is only used to mark the location of the answers for better visualization and is not included in the input image.} 
    \label{fig:table_parse1}
\end{figure*}

\begin{figure*}[tp]
    \centering
    \includegraphics[width=1.0\linewidth]{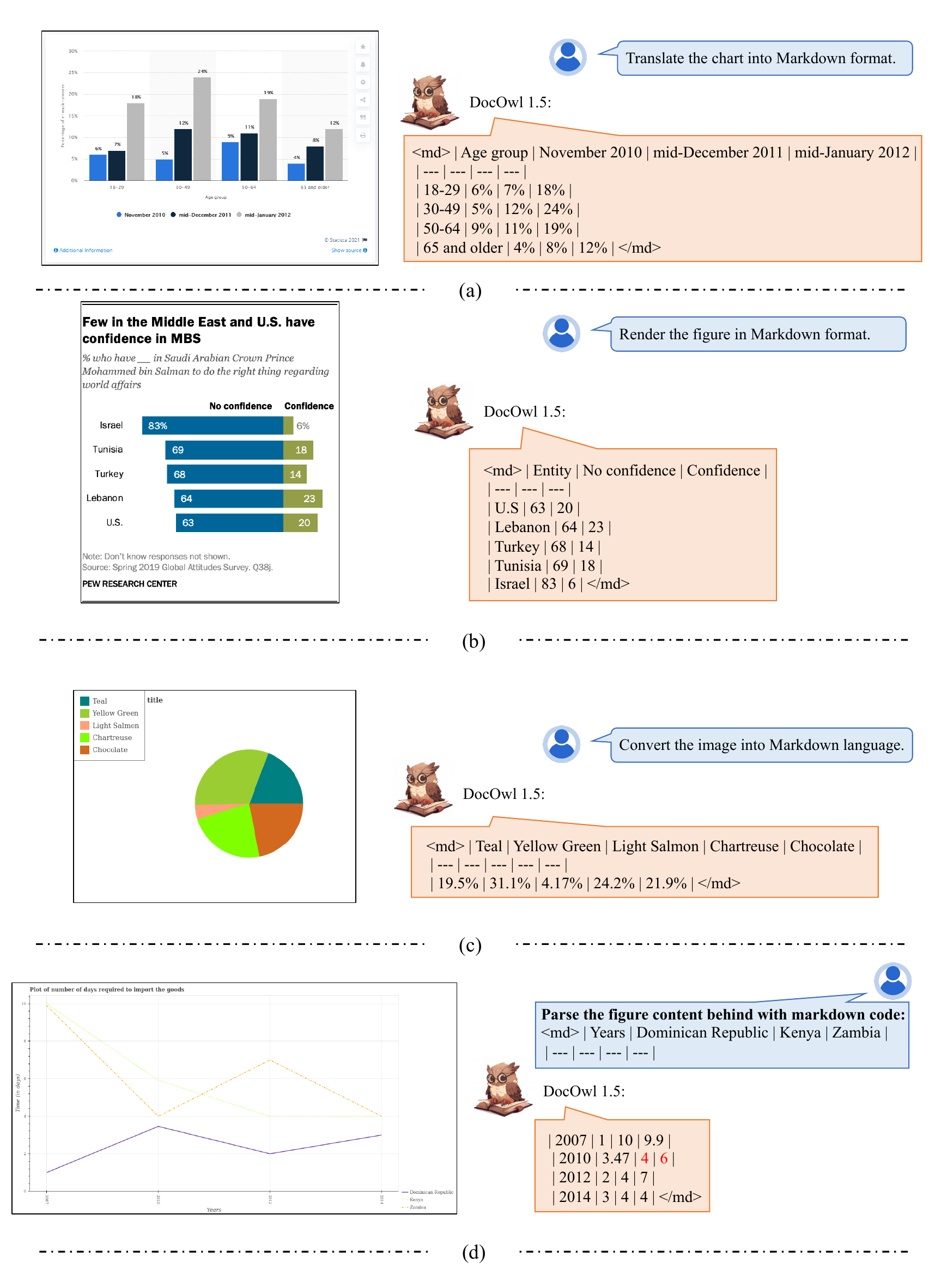}
    \caption{Qualitative results of structure-aware chart parsing on the charts of the vertical bar (a), horizontal bar (b), pie (c), and line (d). Incorrect words in the answer are marked in {\color{red}red}.} 
    \label{fig:chart_parse1}
\end{figure*}

\begin{figure*}[tp]
    \centering
    \includegraphics[width=1.0\linewidth]{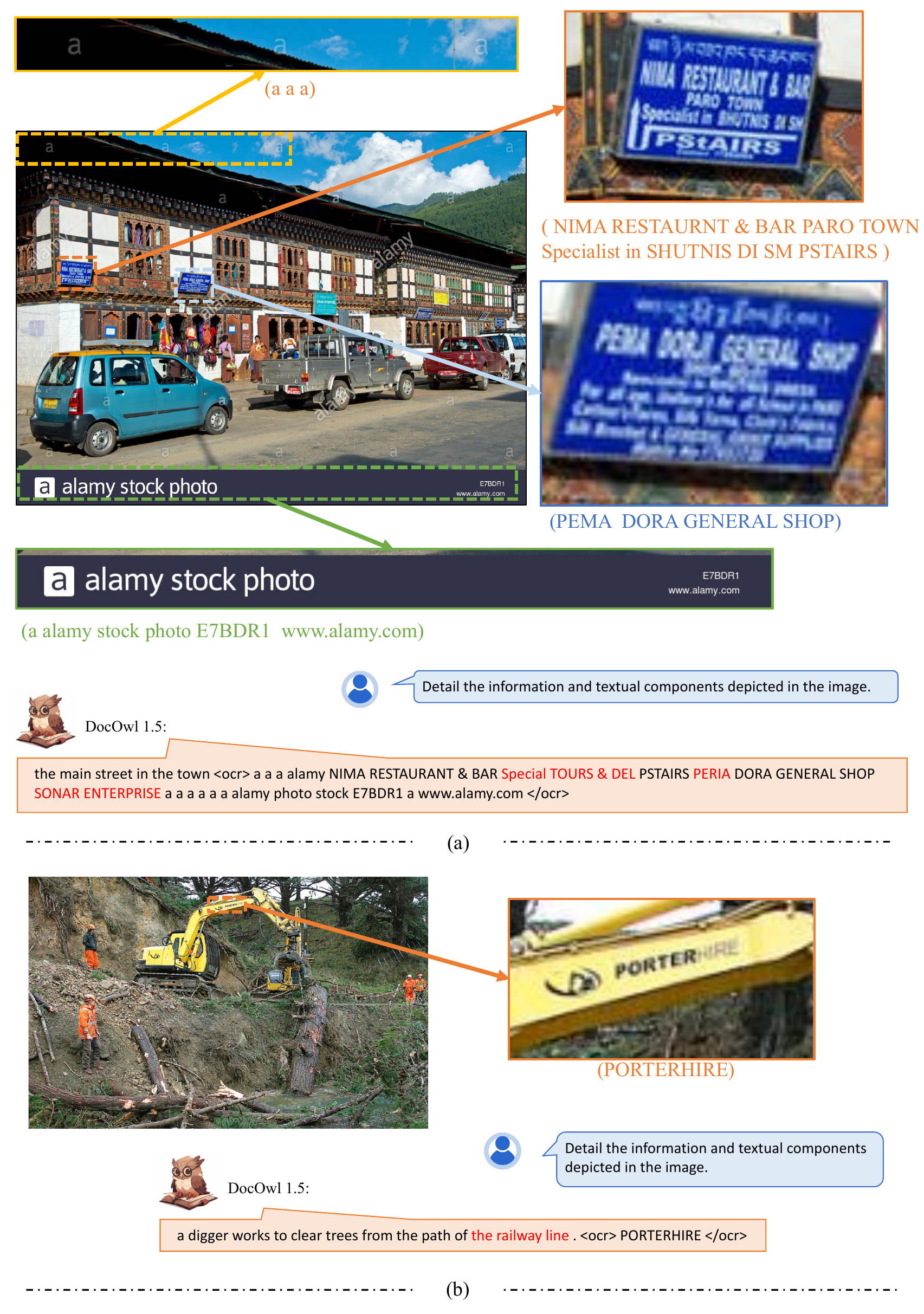}
    \caption{Qualitative results of natural image parsing. For better visualization, some regions are enlarged and labeled with corresponding scene texts. Incorrect words in the answer are marked in {\color{red}red}.} 
    \label{fig:natural_parse1}
\end{figure*}

\noindent\textbf{Structure-aware Parsing.} As shown in \cref{fig:doc_parse}, \modelname~could parse a document image by using line feeds and spaces to represent the structure of text contents. Besides parsing the whole document, as shown in \cref{fig:doc_parse2}, it could also parse texts from the middle of the image according to human instruction. \cref{fig:table_parse1} presents qualitative results of structure-aware table parsing through extended Markdown syntax on tables with cells spanning multiple columns or not. Furthermore, \cref{fig:chart_parse1} shows some cases of parsing different types of charts into Markdown codes, including vertical bar, horizontal bar, pie, and line charts. When all data points are presented in the chart, \modelname~can accurately align statistic objects with corresponding numbers. It makes some mistakes in \cref{fig:chart_parse1}(d) because estimating the concrete numbers is quite challenging when no data points are provided. Finally, as shown in \cref{fig:natural_parse1}, \modelname~can both describe the content of natural images and read scene texts.

\begin{figure*}[tp]
    \centering
    \includegraphics[width=1.0\linewidth]{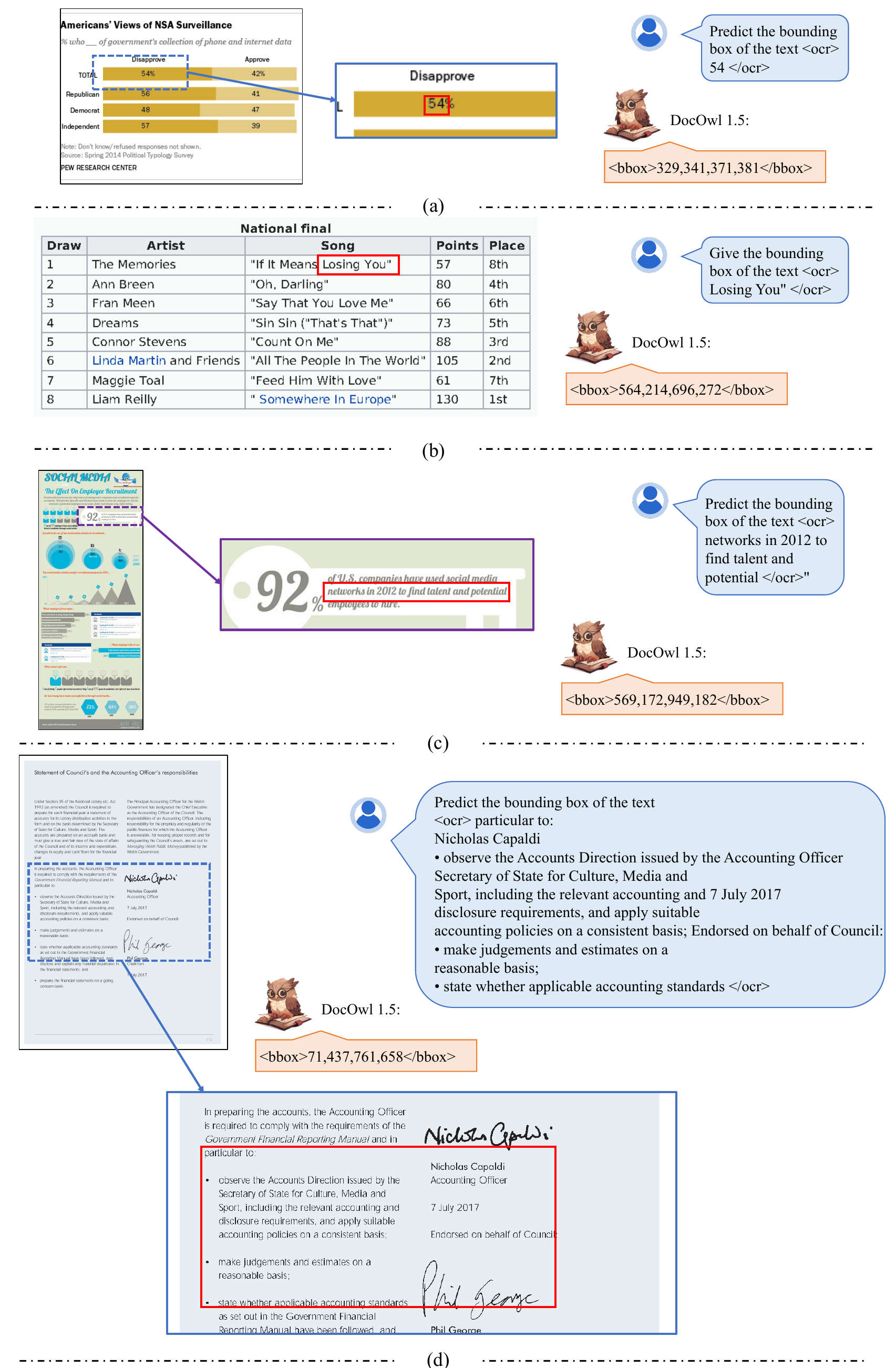}
    \caption{Qualitative results of Multi-grained Text Grounding. Some regions are enlarged for better visualization. Bounding boxes predicted by \modelname~are drawn in images as solid red boxes.} 
    \label{fig:ground}
\end{figure*}

\begin{figure*}[tp]
    \centering
    \includegraphics[width=1.0\linewidth]{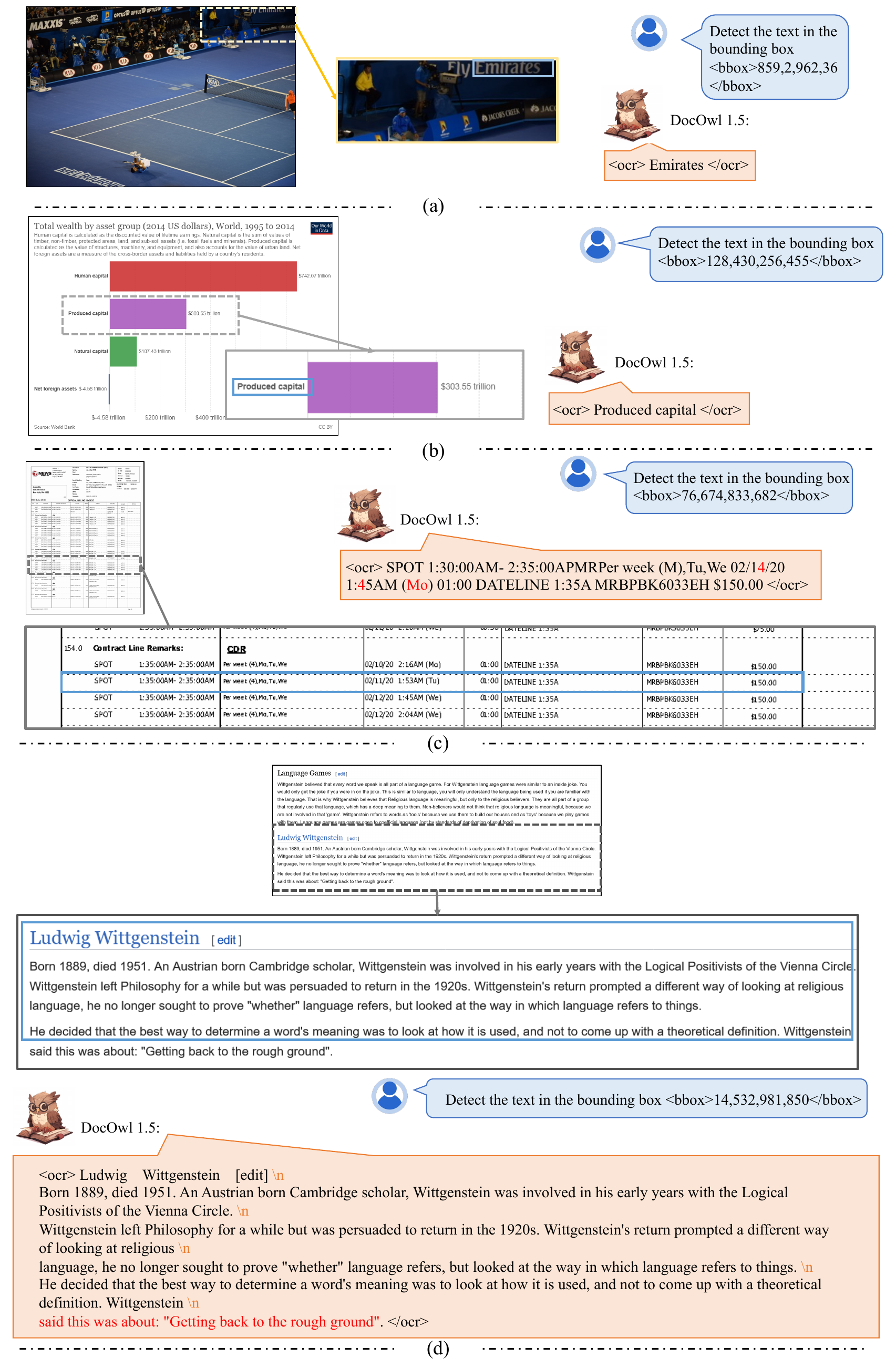}
    \caption{Qualitative results of Multi-grained Text Recognition. Some regions are enlarged for better visualization. Input bounding boxes are drawn in images as solid blue boxes. Incorrect words in answers are marked in {\color{red}red}.} 
    \label{fig:recognize}
\end{figure*}

\noindent\textbf{Multi-grained Text Localization.} \cref{fig:ground} and \cref{fig:recognize} show qualitative results of text grounding and text recognition at granularities of word, phrase, line and block. The image domains range from documents, webpages, charts, and tables to natural images.

\section{Conclusion}
To enhance the Visual Document Understanding performance of Multimodal Large Language Models, we first propose Unified Structure Learning across 5 domains of text-rich images, including both structure-aware parsing tasks and multi-grained text localization tasks. To better maintain structure and spatial information during vision-and-language feature alignment, we design a simple and effective vision-to-text module, named \connectname. It mainly utilizes a convolution layer to aggregate horizontally neighboring visual features. To support the Unified Structure Learning, we build a training dataset \pretraindataname~by collecting publicly available images and carefully constructing structure-aware text sequences and multi-grained pairs of texts and bounding boxes. With Unified Structure Learning, our model \modelname~achieves state-of-the-art OCR-free performance on 10 visual document understanding benchmarks.

\bibliographystyle{plainnat}
\bibliography{main}

\end{document}